\documentclass[journal]{IEEEtran}
\ifCLASSINFOpdf
  \usepackage[pdftex]{graphicx}
\else
  \usepackage[dvipdfmx]{graphicx}
   \DeclareGraphicsExtensions{.pdf}
\fi
\usepackage{booktabs}
\usepackage{comment}
\usepackage{color}
\usepackage[noadjust]{cite}

\usepackage{url}

\begin{document}
%
\title{A Novel Weight-Shared Multi-Stage CNN\\ for Scale Robustness}
%
%
\author{Ryo~Takahashi,
        Takashi~Matsubara,~\IEEEmembership{Member,~IEEE},
        and~Kuniaki~Uehara,
\thanks{R.~Takahashi, T.~Matsubara, and K.~Uehara are with Graduate School of System Informatics, Kobe University, 1-1 Rokko-dai, Nada, Kobe, Hyogo, 657-8501 Japan. E-mails: takahashi@ai.cs.kobe-u.ac.jp, matsubara@phoenix.kobe-u.ac.jp, and uehara@kobe-u.ac.jp.}
}

%
%


\markboth{Journal of \LaTeX\ Class Files,~Vol.~14, No.~8, August~2015}%
{Shell \MakeLowercase{\textit{et al.}}: Bare Demo of IEEEtran.cls for IEEE Journals}
%



\maketitle

\begin{abstract}
Convolutional neural networks (CNNs) have demonstrated remarkable results in image classification for benchmark tasks and practical applications.
The CNNs with deeper architectures have achieved even higher performance recently thanks to their robustness to the parallel shift of objects in images as well as their numerous parameters and the resulting high expression ability.
However, CNNs have a limited robustness to other geometric transformations such as scaling and rotation.
This limits the performance improvement of the deep CNNs, but there is no established solution.
This study focuses on scale transformation and proposes a network architecture called the \emph{weight-shared multi-stage network} (WSMS-Net), which consists of multiple stages of CNNs.
The proposed WSMS-Net is easily combined with existing deep CNNs such as ResNet and DenseNet and enables them to acquire robustness to object scaling.
Experimental results on the CIFAR-10, CIFAR-100, and ImageNet datasets demonstrate that existing deep CNNs combined with the proposed WSMS-Net achieve higher accuracies for image classification tasks with only a minor increase in the number of parameters and computation time.
\end{abstract}


%
\IEEEpeerreviewmaketitle

\section{Introduction}
\label{sec:intro}
Convolutional neural networks (CNNs)~\cite{LeCun1989} have made significant achievements in the tasks of image classification and image processing~\cite{Zeiler2014,Sermanet2014}.
They have already been employed for practical uses in various situations.
CNNs are known to be robust to small parallel shifts thanks to their architecture~\cite{LeCun1989}: Units in a CNN have local receptive fields, share weight parameters with other units, and are sandwiched by pooling layers.
The performance of CNNs has been continuously improved by the development of new network architectures.
A brief history of CNNs can be found in the results of the ImageNet Large Scale Visual Recognition Competition (ILSVRC)~\cite{Russakovsky2014} (e.g., AlexNet~\cite{Krizhevsky2012} in 2012, VGG~\cite{Simonyan2015a} and GoogLeNet~\cite{Szegedy2014} in 2014 and ResNet~\cite{He2016} in 2015).
Especially, the ResNet family is attracting attention thanks to its new shortcut connection that propagates the gradient through more than 100 convolution layers to overcome the gradient vanishing~\cite{Veit2016a} and degradation~\cite{Schmidhuber2015,Bengio1994,Glorot2010} problem, which limited the performance of deep neural networks.

\begin{figure}[!t]
\centering
\includegraphics[width=3.8in,bb= 0 30 820 280,clip]{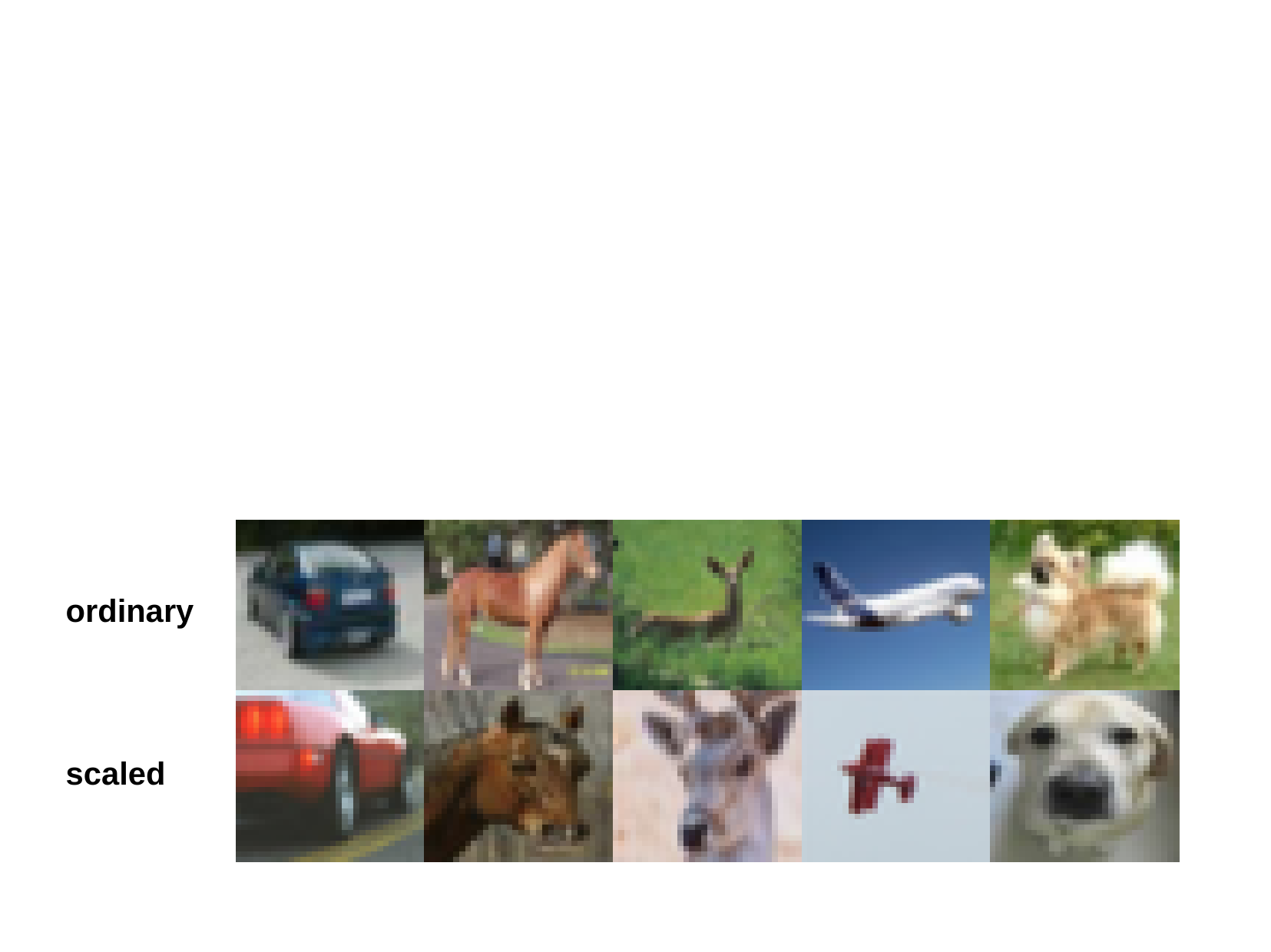}
\caption{
Examples of CIFAR-10 test images showing variability in scale.
The upper row images depict wide-shots and the lower row images depict close-ups.
}
\label{fig:scaled_cifar}
\end{figure}

\begin{figure}[!t]
\centering
\includegraphics[width=3.8in,bb= 0 0 820 600,clip]{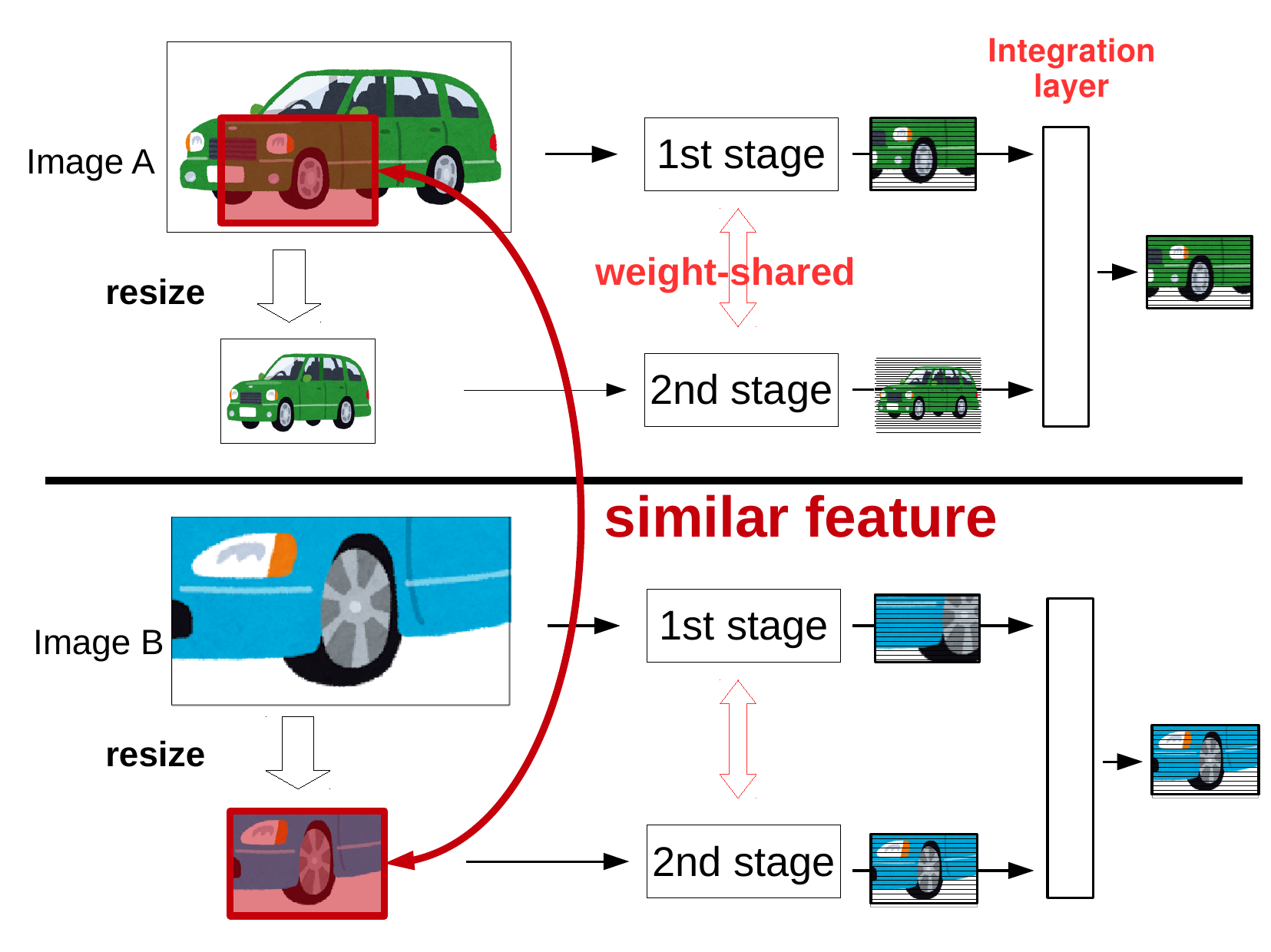}
\caption{Conceptual explanation of multi-scale feature fusion in the proposed \emph{weight-shared multi-stage network} (WSMS-Net).
Image A provides the local features of an automobile such as the shapes of the tires, windows, headlights, and license plate in the first stage.
Similar features are detected in the resized version of the image B, a part of a sports car, in the second stage thanks to the shared weight parameters.}
\label{fig:scale}
\end{figure}

In contrast, CNNs still have limited robustness to geometric transformations other than parallel shifts, such as scaling and rotation~\cite{Le2010}, and this problem has no established solution.
Scaled and rotated objects in images are recognized incorrectly by CNNs or at least require additional parameters recognizing the original objects, limiting the expression ability of CNNs.
Except for a few studies using special datasets~\cite{Jaderberg2015,Gregor2015}, scaling has not been focused on in the classification task.
However, popular image classification benchmark datasets, such as CIFAR~\cite{Krizhevsky2009} and ImageNet~\cite{Russakovsky2014}, contain many close-ups and wide-shots, and thus, dealing with scaling of objects is necessary in the classification task.
Figure~\ref{fig:scaled_cifar} shows close-ups and wide-shots in the CIFAR-10 (lower row) as well as other ordinary images (upper row).
How the scaling is dealt with is important for model's performance.
On the other hand, in segmentation and object detection tasks, many previous works tackled scaling of objects using multi-scale feature extraction~\cite{Cai2016,Bargoti2017,Audebert2016,Marmanis2016a}.
This is because performance greatly depends on the detection of the object region in an image robustly to its scale.
Unfortunately, the focuses of these works were mainly on segmentation tasks, but not on classification task.
These works cannot be directly applied to classification task.
We demonstrate this issue in Section~\ref{sec:previous_work} and propose a new network architecture robust to scaling of objects in classification task in Section~\ref{sec:propose}.

In this paper, we propose a network architecture called the \emph{weight-shared multi-stage network} (WSMS-Net).
In contrast to ordinary CNNs, which are built by stacking convolution layers, a WSMS-Net has a multi-stage architecture consisting of multiple ordinary CNNs arranged in parallel.
A conceptual diagram of the WSMS-Net is depicted in Fig.~\ref{fig:scale}.
Each stage of the WSMS-Net consists of all or part of the same CNN: The weights of each convolution layer in each stage are shared with those of the corresponding layers of the other stages.
Each stage is given a scaled image of the original input image.
The features extracted at all stages are concatenated and fed to the \emph{integration layer}.
We expect that the integration layer selects the feature for classification and discards the remaining information.
If the feature from one of the stages is sufficient for classification, the integration layer selects it, and the other stages require no further training.
At the Fig.~\ref{fig:scale}, in the case of inputting image B after learning the image A, the integration layer is expected to select the feature from stage $2$ similar to the feature learned from image A.
Thanks to this architecture, similar features are extracted even from objects of different scales at the various stages, and thereby, the same objects of different scales are classified into the same class.
We apply the WSMS-Net to existing deep CNNs and evaluate them on the CIFAR-10, CIFAR-100~\cite{Krizhevsky2009} and ImageNet~\cite{Russakovsky2014} datasets.
The experimental results demonstrate that the WSMS-Nets significantly make the original deep CNNs robust to scaling and improve the classification accuracy.
A preliminary and limited result is found in a conference paper~\cite{Takahashi2017b}.

\section{Related Works}
\label{sec:Related_works}

\subsection{Deep CNNs}
CNNs have been introduced with shared weight parameters, limited receptive fields, and pooling layers~\cite{LeCun1989}, inspired by the biological architecture of the mammalian visual cortex~\cite{Hubel1968,Fukushima1980}.
Thanks to this architecture, CNNs can suppress increases in the number of weight parameters and are robust to the parallel shift of objects in images.
In recent years, CNNs have made remarkable improvements in image classification with increasingly deeper architectures.
In general, CNNs use the back-propagation algorithm~\cite{LeCun1989}, which calculates the gradient obtained at the output layer, back-propagates the gradient from the output layer to the input layer, and updates the weight parameters.
However, when the network becomes very deep, the gradient information from the output layer does not pass well to the input layer and other layers near the input, and learning does not progress.

\subsection{State-of-the-Art Deep CNNs for Image Classification}
To address the gradient vanishing~\cite{Veit2016a} and degradation~\cite{Schmidhuber2015,Bengio1994,Glorot2010} problem, ResNet~\cite{He2016} and DenseNet~\cite{Huang2016b} were proposed as network architectures that enable learning with even deep architecture for image classification.
We introduce these two works because we use them as the basic works for our WSMS-Net method.
ResNet is the network that recorded the highest accuracy in the ILSVRC image classification task in 2015.
Following ResNet, many ideas and new architectures have been proposed~\cite{He2016b,Zagoruyko2016,Huang2016a,Targ2016a,Han2016}, and in ILSVRC 2016, an extended version of ResNet broke the record~\cite{Zagoruyko2016}.

ResNet (Residual Network)~\cite{He2016} is a new architecture that enables the CNNs to overcome the gradient vanishing and degradation problem.
The basic structure of ResNet is called a \emph{residual block}, which is composed of convolution layers and a shortcut connection that does not pass through the convolution layers: The shortcut connection simply performs identity mapping.
The result obtained from the usual convolution layers is denoted by $F(x)$, and the shortcut connection output is denoted by $x$.
The output obtained from the whole block is $F(x)+x$.
In deep CNNs, the gradient information can vanish at the feature extraction point in the convolution layer.
However, by adding an identity mapping, the gradient information can be transmitted without any risk of gradient vanishing and degradation.
The whole network of ResNet is built by repeatedly stacking residual blocks.
\emph{Residual block} needs no extra parameters, and the gradient can be calculated in the same manner as conventional CNNs.
ResNet employs a residual block consisting of a shortcut and two convolution layers of $3\times3$ kernels for a fixed number of the channels.
In the original study~\cite{He2016}, each layer is constructed using a conv-BN-ReLU order.
In a later study~\cite{He2016b}, a modified ResNet was proposed with layers constructed using a BN-ReLU-conv order that achieved a higher classification accuracy.
BN indicates the batch normalization~\cite{Ioffe2015} here.
This modified ResNet is called Pre-ResNet.

DenseNet (Densely Connected Convolutional Network)~\cite{Huang2016b} is a network that improves upon concept of ResNet~\cite{He2016}.
Its image classification accuracy is superior to that of ResNet.
As its name implies, the structure of DenseNet connects layers densely: In contrast to ResNet, in which a shortcut is connected across two convolution layers, a shortcut is connected from one layer to \emph{all} subsequent layers in DenseNet.
In addition, the output of the shortcut is not added to but is concatenated with the output of a subsequent convolution layer in the channel direction of the feature map.
The number of channels of the feature map increases linearly as the network is deepened, and the number $k$ of the increased channels per layer is called the \emph{growth rate}.
The basic architecture of DenseNet is called a \emph{dense block}, and the whole network of a DenseNet is built by stacking this dense blocks.
The original DenseNet consists of three dense blocks: The size of the feature map in a single dense block is fixed, and the shortcuts are not connected across each dense block.

Although the development of new network models such as ResNet and DenseNet has overcome the gradient vanishing and degradation problem, CNNs still do not have an established solution to geometric transformations such as scaling and rotation.
Lack of invariance to these transformations is an obstacle to the progress of deep CNNs.

\subsection{CNNs against Geometric Transformations}
\label{sec:related_c}
A few studies in classification task have tried to address geometric transformations using neural networks~\cite{Jaderberg2015,Gregor2015}.
The Spatial Transformer Network (STN) and Deep Recurrent Attentive Writer (DRAW) infer parameters such as position and the angle of the geometric transformation of an object in an image.
They are general and powerful approaches to make a network robust against geometric transformations.
However, the STN requires additional CNNs to localize and correct an object in an image.
DRAW requires repeated computations between all-to-all pairs of pixels of the input and output images to adjust parameters gradually for each image.
They require many additional parameters and computation time, and thereby, not suitable for combining with deep CNNs.
These works can adapt to arbitrary scales if trained with images of appropriate scales.
They are not generalized to unknown scales; they are not scale-invariant strictly.
Other studies \cite{Lowe2002,Lowe2004,Koenderink1984,Lindeberg1993} deal with the scale-invariance in the strict definition but are unavailable with neural networks.

In contrast, many works focused on geometric transformations in segmentation and object detection tasks.
Cai et al.~\cite{Cai2016} aimed fast multi-scale object detection on the KITTI~\cite{Geiger2012} and Caltech~\cite{Wojek2012} datasets, which contain a substantial number of the small objects.
This work proposed a unified multi-scale deep CNN, which has multiple output layers so that their receptive fields catch objects of different scales.
Feature maps of different sizes are acquired from different output layers and used like an ensemble.
Bargoti et al.~\cite{Bargoti2017} proposed a network dealing with multiple input images.
Input images of different sizes are input into the network, and their features are concatenated to a feature map in the next layer.
This is called multi-scale input patch.
Audebert et al.~\cite{Audebert2016} proposed a multi-kernel convolution that operates at three multiple scales.
An image is given to three convolution layers with $3\times3$, $5\times5$, and $7\times7$ kernels in parallel, and their outputs are concatenated to one feature map.
Marmanis et al.~\cite{Marmanis2016a} used a simple ensemble of the outputs of a CNN given an image resized to multiple scales.
These works are also not strictly scale-invariant but are robust to scaling.

\section{Weight-Shared Multi-Stage Network}
\subsection{Architecture of Weight-Shared Multi-Stage Network}
\label{sec:propose}

\begin{figure}[!t]
\centering
\includegraphics[width=3.6in,bb= 0 40 820 550,clip]{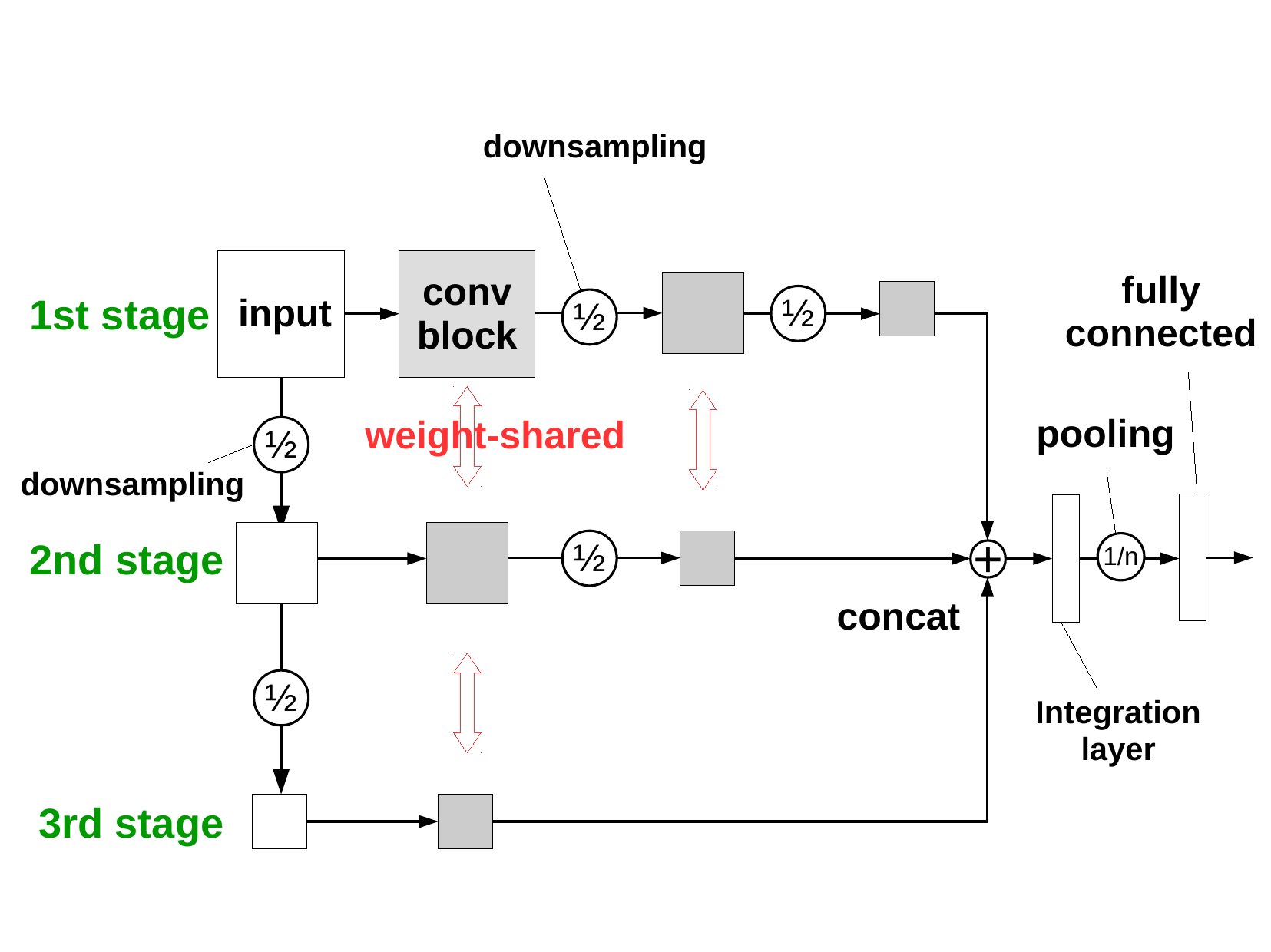}
\caption{A basic WSMS-Net consisting of three stages.
The input image is resized to half and a quarter for the second stage and third stages, respectively.
Each \emph{conv block} consists of some convolution layers.
The \emph{weight-shared} label indicates that the weight parameters of all the convolution layers in the conv blocks in multiple stages are completely shared with each other.
After each conv block, the feature map is resized to half at a \emph{downsampling} layer, which is typically an average pooling, a max pooling, or a convolution layer with a stride of 2 but depends on the original CNNs.
The output feature maps of all the stages are concatenated in the channel direction at the circle indicated by \emph{concat}.}
\label{fig:MSNet_related_work}
\end{figure}

In this study, we propose a novel network architecture called \emph{weight-shared multi-stage network} (WSMS-Net) for acquiring robustness to object scaling.
The basic architecture of WSMS-Net is shown in Fig.~\ref{fig:MSNet_related_work}.
A WSMS-Net consists of $S$ stages, and each stage is either the entirety or part of an ordinary CNN that can be divided into $t$ convolution blocks for $t\ge S$.
Each convolution block consists of some convolution layers.
After each convolution block, the feature map is resized to half at a \emph{downsampling} layer.
Each \emph{conv block} consists of some convolution layers.
We apply this WSMS-Net architecture to existing deep CNNs; ResNet and DenseNet.
A conv block corresponds to a dense block of DenseNet and a sequence of \emph{residual blocks} of ResNet.
A downsampling is performed at a pooling or convolution layer with a stride of 2.
Note that, for WSMS-Net, we replaced a convolution layer with a stride of 2 in ResNet with a convolution layer with a stride of 1 followed by a pooling layer with a stride of 2 for a robust result.
The first stage consists of all $t$ blocks while the second stage has the first $(t-1)$ blocks, the third stage has the first $(t-2)$ blocks, and so on.
The convolution blocks at the same depth of all the stages completely share the weight parameters of the convolution layers.
Note that when batch normalization~\cite{Ioffe2015} is employed, its parameters are not shared.
The first stage is given the original image, while the second stage is given the image resized by half by a downsampling layer implemented as an average pooling layer.
Note that an average pooling with a kernel size of 2 and a stride of 2 is equivalent to an interpolation for resizing an image to $\frac{1}{2}$.
The WSMS-Net only resizes an image to $\frac{1}{2}$ and implements this operation as an average pooling.
Similarly, the $s$-th stage is given the image resized to half of the size of the image given to the ($s\!-\!1$)-th stage.
Therefore, the sizes of the output feature maps of all the stages are the same.
These feature maps are then concatenated in the channel direction at the ends of all the stages.
After concatenation, all of the feature maps are integrated at the integration layer and are used as input to the last fully connected layer for classification.
Thanks to these processes, various features are extracted from the input image at multiple sizes in multiple stages and hence contribute to the robustness to scaling.

\begin{figure}[!t]
\centering
\includegraphics[width=3.7in,bb= 0 00 820 650,clip]{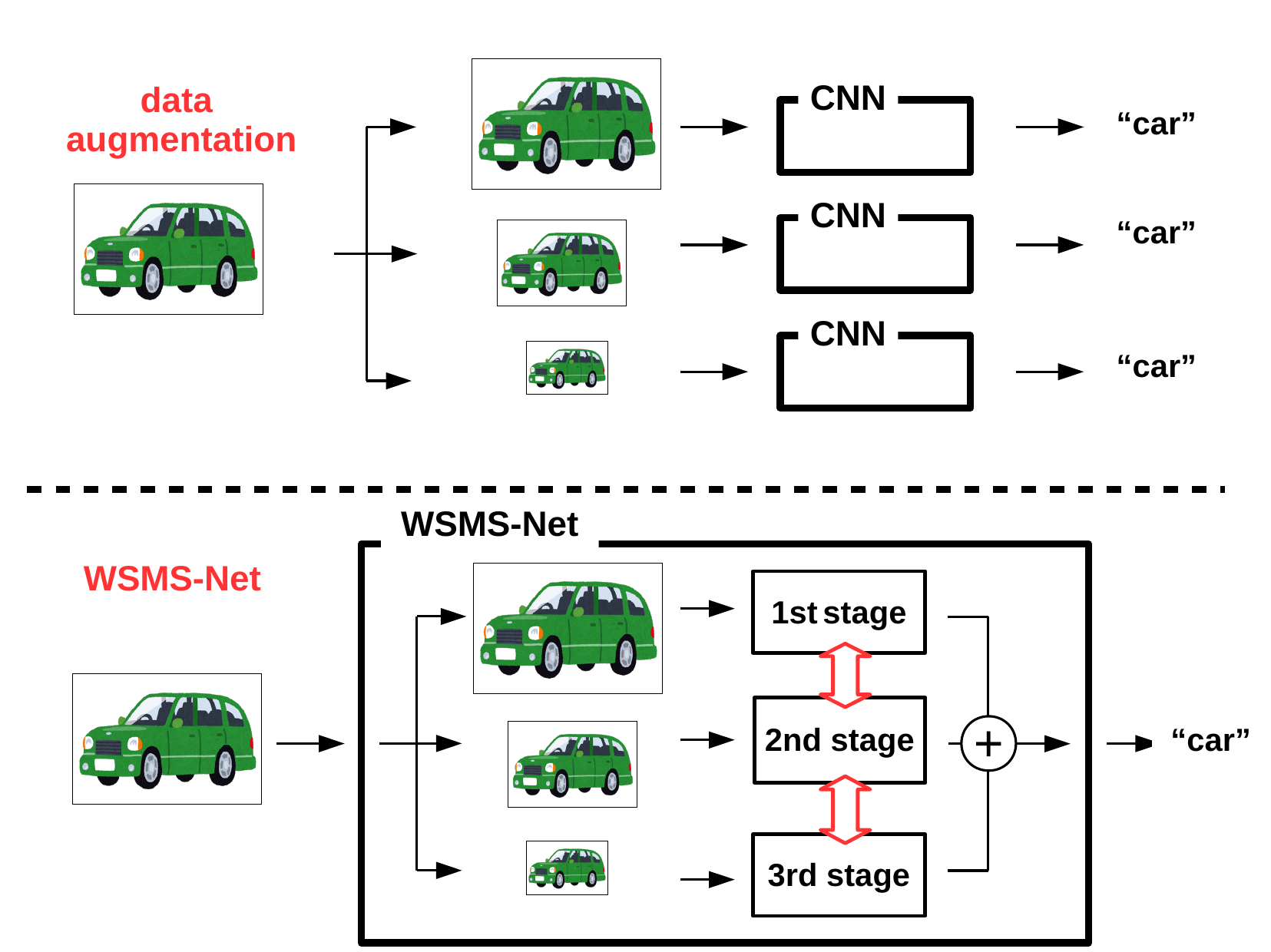}
\caption{Difference between data augmentation and WSMS-Net.
With data augmentation, CNN needs numerous parameters to learn the various features from resized images.
An input image is resized in WSMS-Net consisting of several stages, and WSMS-Net learns at least one significant feature from the input images of multiple sizes by weight sharing.
Therefore, WSMS-Net does not need extra parameters.
}
\label{fig:aug_wsms_dif}
\end{figure}

\subsection{Concepts of Weight-Shared Multi-Stage Network}
\label{sec:concept}
We explain the detailed concept of a WSMS-Net in this section.
Consider images A and B, depicted in Fig.~\ref{fig:scale}.
There are two stages: The first stage is given the original $N \times N$ image as input and the second stage is given the $\frac{N}{2}\times\frac{N}{2}$ image resized by half as input.
Image A shows the entire automobile and image B shows only the part around the left front tire of a sports car.
In the training phase, WSMS-Net is given the image A and learns the features related to the automobile.
In the first stage, because the size of the image remains at $N\times N$, WSMS-Net learns the local features of the car such as the shapes of the tires, windows, headlights, and license plate.
In the second stage, because the image is resized to $\frac{N}{2}$, WSMS-Net learns the features such as the whole shape of the car but ignores the local features learned in the first stage.
Between the two stages, different features are found: This is necessary for our WSMS-Net to become robust to scaling of objects.
We expect that a wide-shot and a close-up of similar objects have similar features in the first and second stages.
With image A, WSMS-Net learns the feature maps in the red shaded area see Fig.~\ref{fig:scale}, in its first stage.
After that, WSMS-Net finds similar feature maps in the image B in the second stage.
Thanks to the different influences of receptive fields of two stages, feature maps in the red shaded area in Fig.~\ref{fig:scale} in the first stage learned by the image A are similar to those in the second stage learned by the image B.
If images A and B are given to an ordinary CNN, the CNN cannot find any features shared by them.

We emphasize that WSMS-Net is not the same as data augmentation~\cite{Lee2014}.
The difference between data augmentation and WSMS-Net is depicted in Fig.~\ref{fig:aug_wsms_dif}.
With data augmentation, CNNs are always given a single image resized with various scaling factors, and they try to classify it without other clues.
To classify an object consistently across different scales, a CNN learns the same feature of different scales as different features (e.g., features of the image A and image B in Fig.~\ref{fig:scale} are completely different).
During the training procedure with data augmentation method, the CNN learns features (e.g., a front tire) from a scaled-up version of the image A and classify the image B using that feature since they are similar to each other after scaling (see the red areas).
However, the CNN also must learn other features from the image A of the default size, from a scaled-up version of the image B, and a scaled-down version of the image A.
In short, the CNN has to learn a wide variety of features using much more weight parameters.
On the contrary, WSMS-Net utilizes the same weight parameters responding to similar features for classifying both the image A and image B.
The integration layer receives features of multiple scales from multiple stages.
If the feature from one of the stages is sufficient for classification, the integration layer selects it, and the other stages require no further training.
As a result, the variety of the features that WSMS-Net has to learn is reduced.
A WSMS-Net can be considered to share weight parameters in the front-back direction of images in addition to the ordinary CNNs sharing weight parameters only in the vertical and horizontal directions.

\section{Optimization of WSMS-Net}
\label{sec:opt}

\subsection{Combination with Existing CNNs}
The combining of ResNet, Pre-ResNet, and DenseNet with the WSMS-Net(called WSMS-ResNet, WSMS-Pre-ResNet, and WSMS-DenseNet respectively) should enable them to classify the images that the original CNNs could not.
The number of stages and shape of the integration layer need to be optimized to construct WSMS-ResNet, WSMS-Pre-ResNet, and WSMS-DenseNet.
The integration layer is an extra layer placed just after the concatenation layer that integrates all the feature maps before the classification.
For comparison, we also introduce a more trivial network called the \emph{multi-stage network} (MS-Net).
An MS-Net has its own multiple stages like a WSMS-Net, but each stage has weight parameters, similar to an ensemble of multiple CNNs (e.g., \cite{Zhang2016a}).
MS-Net thus has much more weight parameters than WSMS-Net but does not have the aforementioned features of WSMS-Net.

\subsection{Numbers of Stages}

The number of stages is a fundamental factor of WSMS-Net.
Typical CNNs can be divided into several compartments by pooling layers (e.g., an average pooling layer~\cite{LeCun1989} or max pooling layer~\cite{Riesenhuber1999}), which downsampled the feature maps: ResNet and DenseNet employ a convolution layer followed by an average pooling layer.
Each compartment of ResNet consists of several residual blocks and each compartment of DenseNet consists of a single dense block.
In this paper, one compartment is treated as a single convolution block of WSMS-Net.
As a result, WSMS-ResNet, WSMS-Pre-ResNet, and WSMS-DenseNet can have the stages as many as or less than the compartments.

\begin{figure}[!t]
\centering
\includegraphics[width=3.6in,bb= 30 0 708 600,clip]{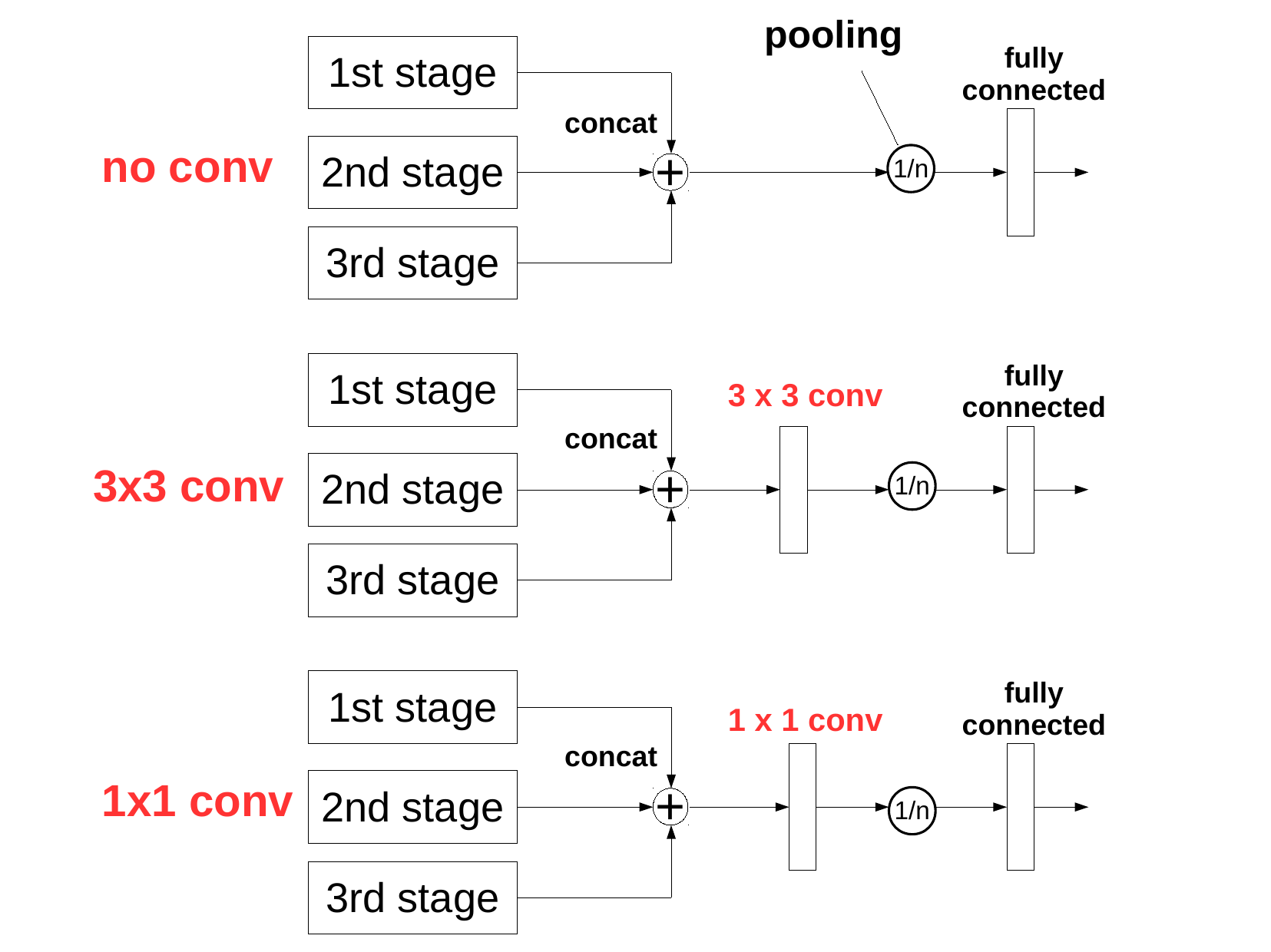}
\caption{Three types of integration layers.
With \emph{no conv}, the concatenated feature map is given directly to the global pooling layer followed by the last fully connected layer.
Moreover, $3\times3$ \emph{conv} and $1\times1$ \emph{conv} are $3\times3$ and $1\times1$ convolution layers, placed between the concatenated feature map and the global pooling layer.
}
\label{fig:Integration}
\end{figure}

\subsection{Integration Layer}
\label{sec:integration}

We have not yet described the integration layer in detail.
The feature map obtained from stacked convolution layers is often given to a global pooling layer and becomes a feature vector before the last fully connected layer, whose outputs denote the inferred class labels.
Owing to this concatenation, the final number of channels of the feature map is larger than that of the original network.
To arrange the multiple features from multiple stages in WSMS-Net, an integration layer needs to select the most useful features.
We explain this selection by the integration layer in the case of WSMS-Net.
For example, an image depicts an $S$ size object $X$ and other image depicts a $2S$ size object $X$.
Given these two images, after the first to third stages, the integration layer gets the concatenated feature vectors ($S$, $S/2$, $S/4$) and ($2S$, $S$, $S/2$).
Hence, these feature vectors are not the same.
However, the integration layer in addition to the last fully connected layer forms the so-called multilayer perceptron (mimicking an arbitrary function) and is expected to select a salient feature from those extracted by the multiple stages.
Then, we expect the integration layer to select the common feature ($S$, $S/2$) from the feature vectors.

We consider several candidates for an integration layer, as shown in Fig.~\ref{fig:Integration}.
In the simplest way, the integration layer does nothing, and the last fully connected layer is given a feature vector longer than that of the original network.
Then, the last fully connected layer linearly sums the feature vector like an ensemble method, which is commonly used for arranging multiple outputs from a network.
This integration layer is called \emph{no conv} hereafter.
On the contrary to no conv, we can use a convolution layer as an integration layer.
This extra convolution layer, in addition to the last fully connected layer, forms so-called multilayer perceptron (mimicking an arbitrary function) and is expected to  select a salient feature from those extracted by the multiple stages.
We evaluated a $3\times 3$ convolution layer as an integration layer because both ResNet and DenseNet mainly employ $3\times 3$ convolution layers.
This integration layer is called $3\times 3$ \emph{conv}.
We also evaluated a $1\times 1$ convolution layer, called $1\times 1$ \emph{conv}, as an integration layer.
$1\times1$ conv is the simplest convolution layer, and suppresses an increase in the number of weight parameters.
In addition, $1\times1$ conv integration layer does not perform spatial convolution.
Hence, $1\times1$ conv can evaluate the necessity of spatial convolution for feature selection at the integration layer.

\section{Experiments}
\label{chp:exp}


\subsection{Classification of CIFAR-10 and CIFAR-100 by WSMS-DenseNet}
\label{sec:wsms-densenet}
\paragraph*{\textbf{WSMS-DenseNet on the CIFAR-10 and CIFAR-100}}

In this section, we combined a DenseNet with a growth rate $k=24$ with WSMS-Net to construct a WSMS-DenseNet, and evaluated the WSMS-DenseNet on the CIFAR-10 and CIFAR-100 datasets~\cite{Krizhevsky2009}.
CIFAR-10 and CIFAR-100 consist $32\times32$ RGB images of natural scene objects; Each consists of 50,000 training images and 10,000 test images.
Each image is manually given one of 10 class labels in CIFAR-10, and 100 class labels in CIFAR-100 (The number of images per class is thus reduced in CIFAR-100).
DenseNet ($k=24$) was reported to be the model that achieved the second highest classification accuracy among the results of the CIFAR-10 and CIFAR-100 datasets in the original study~\cite{Huang2016b}.
The original DenseNet ($k=24$) has three dense blocks, and the sizes of the feature map are $32\times32$, $16\times16$, and $8\times8$ in the first, second, and third dense blocks, respectively.
There are $3$ channels at the input, which is increased to 16 by the convolution layer placed before the first dense block, and the amount of channel is further increased by the growth rate $k=24$ at every convolution layer.
Each dense block consists of 32 convolution layers.
Therefore, the total number of channels is $16+24\times32\times3=2,320$.
After the third dense block, global pooling is performed, resulting in a 2,320-dimensional feature vector then given to the last fully connected layer for classification.

\begin{table*}[!t]
\renewcommand{\arraystretch}{1.3}
\caption{Test Error Rates of the WSMS-DenseNet and Original DenseNet on the CIFAR-10 and CIFAR-100 Datasets.}
\label{tab:WSMS-DenseNet_CIFAR-10}
\centering
\scalebox{1.0}{
\begin{tabular}{lccccc}
  \toprule
    & & \multicolumn{2}{c}{C10+} & \multicolumn{2}{c}{C100+} \\
  \cmidrule(l{.75em}r{.75em}){3-4}
  \cmidrule(l{.75em}r{.75em}){5-6}
    Network                           & growth rate $k$ & \#params & Error (\%) & \#params & Error (\%) \\
  \midrule
    DenseNet                          & 24              & 27.2M    & 3.74$^\ast$ & 27.2M &  19.25$^\ast$ \\
    DenseNet                          & 26              & 31.9M    & 3.82 & 31.9M & \textbf{18.94} \\
    MS-DenseNet  ($3$ stages, $1\times1$ conv) (not weight-shared for comparison)     & 24 & 41.3M    & 4.18 & 41.3M & \textbf{18.70} \\
  \midrule
    WSMS-DenseNet ($2$ stages, no conv)           & 24              & 27.4M    & 3.98 & 27.7M & 19.68 \\
    WSMS-DenseNet ($2$ stages, $3\times3$ conv)   & 24              & 31.8M    & \textbf{3.54} & 31.8M & \textbf{18.75} \\
    WSMS-DenseNet ($2$ stages, $1\times1$ conv)   & 24              & 27.8M    & \textbf{3.58} & 27.8M  & 19.28 \\
  \midrule
    WSMS-DenseNet ($3$ stages, no conv)           & 24              & 27.4M    & 4.54 & 27.8M & 20.11 \\
    WSMS-DenseNet ($3$ stages, $3\times3$ conv)   & 24              & 32.7M    & \textbf{3.54} & 32.7M & \textbf{19.16} \\
    \textbf{WSMS-DenseNet ($3$ stages, $1\times1$ conv)}   & 24              & 28.0M    & \textcolor{blue}{\textbf{3.51}} & 28.0M  & \textcolor{blue}{\textbf{18.45}} \\
  \bottomrule
  \\
  \multicolumn{3}{l}{$^\ast$ indicates the result cited from original study~\cite{Huang2016b}.}
\end{tabular}
}
\end{table*}

\begin{figure}[!t]
\centering
\includegraphics[width=3.6in,bb= 0 0 820 600,clip]{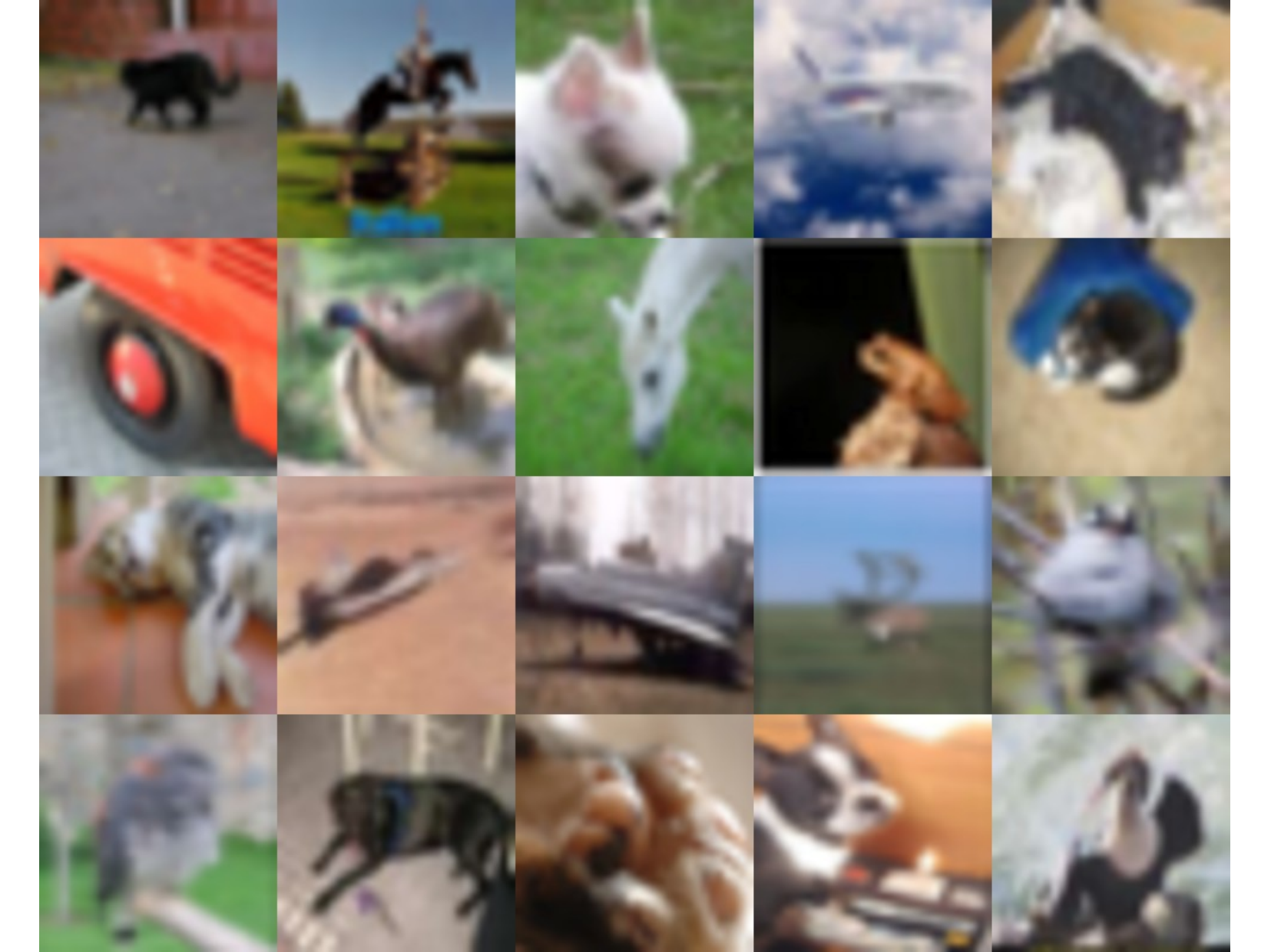}
\caption{Examples of CIFAR-10 test images misclassified by the DenseNet ($k=24$) and the DenseNet ($k=26$) but classified by the WSMS-DenseNet ($k=24$, $1\times1$ conv) correctly.}
\label{fig:ScaleResult24}
\vspace*{1cm}
\centering
\includegraphics[width=3.6in,bb= 0 0 820 600,clip]{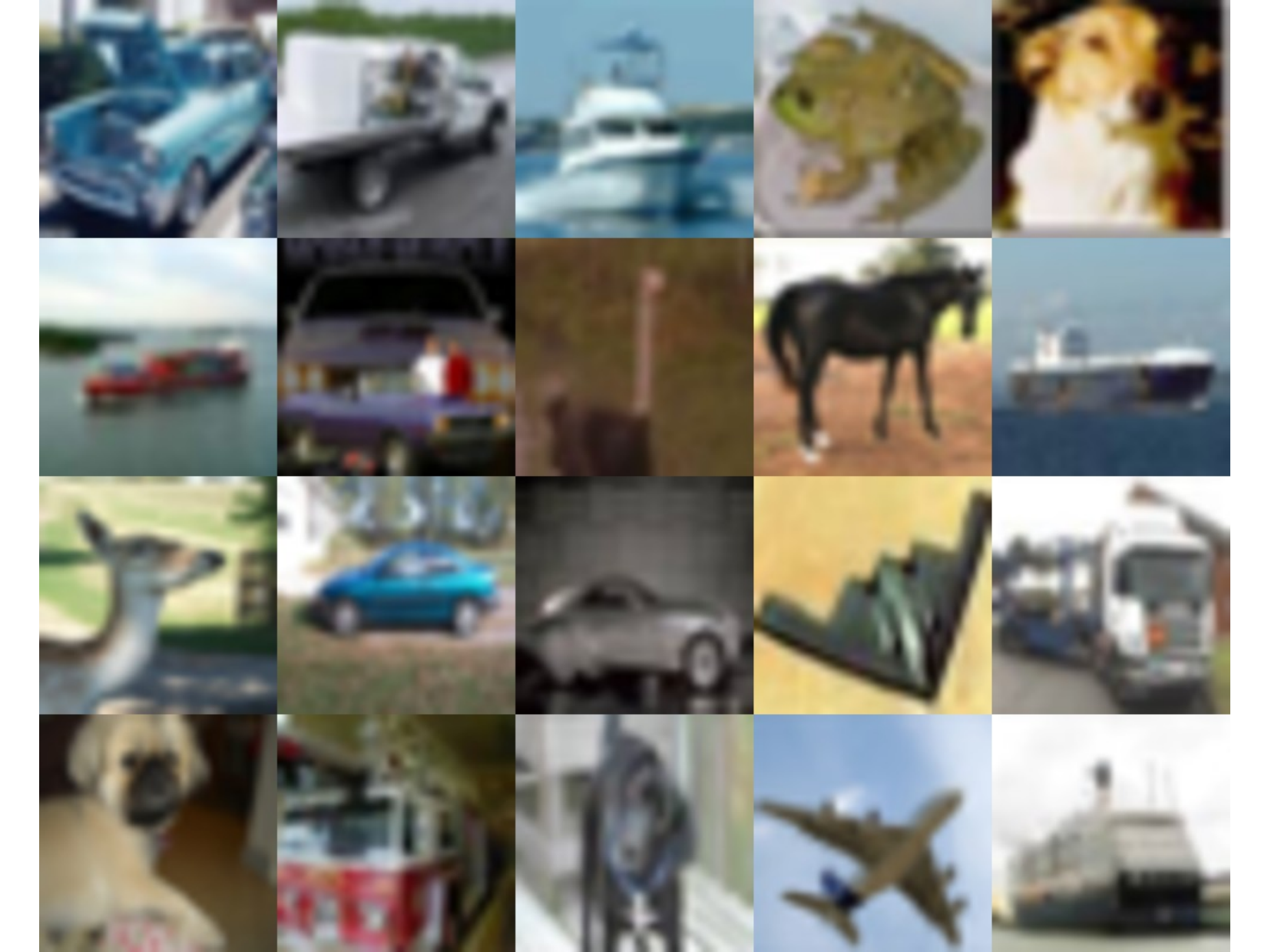}
\caption{Examples of CIFAR-10 test images.}
\label{fig:CIFAR10TEST}
\end{figure}

We constructed WSMS-DenseNet with a growth rate of $k=24$ by combining the DenseNet with WSMS-Net.
Each \emph{dense block} of the DenseNet was treated as a convolution block of the WSMS-Net and we set the number of stages to two patterns of two and three.
The final numbers of channels for the first, second, and third stages were $16+24\times32\times3=2,320$, $16+24\times32\times2=1,552$, and $16+24\times32\times1=784$, respectively.
Hence, the integration layer was given a feature map of $2,320+1,552+784=4,656$ channels.
With the no conv integration layer, the feature maps became a 4,656-dimensional feature vector through the global pooling layer.
We set $c$, the number of the channels of the final feature map, to 128: The feature map was projected to $8\times8$ feature maps of 128 channels through the $3\times3$ conv or $1\times1$ conv integration layer before the global pooling layer.
The hyperparameters and other structural parameters for WSMS-DenseNet followed those of the original DenseNet ($k=24$).
All the images were normalized with the mean and variance of each filter, and a $4$-pixel padding on each side, $32\times32$ cropping, and random flipping in the horizontal direction were employed as further data normalization and data augmentation~\cite{Lee2014,Romero2015,Springenberg2015}.
Batch normalization~\cite{Ioffe2015} and the ReLU activation function~\cite{Nair2010} were used.
The weight parameters were initialized following the algorithm proposed in ~\cite{He2016a}.
WSMS-DenseNet was trained using the momentum SGD algorithm with a momentum parameter of 0.9 and weight decay of $10^{-4}$ over 300 epochs.
The learning rate was initialized to 0.1, and then, it was reduced to 0.01 and 0.001 at the 150th and the 225th epochs respectively.
Note that, the mini-batch size was changed from 64 to 32 owing to the capacity of the computer we used; This change has could potentially degrade the classification accuracy of WSMS-DenseNet.

\paragraph*{\textbf{Classification Results}}

Table~\ref{tab:WSMS-DenseNet_CIFAR-10} summarizes the results of the WSMS-DenseNet ($k=24$) and original DenseNet.
We focus on the results of WSMS-DenseNet with $3$ stages first.
For CIFAR-10, the error rate of our proposed WSMS-DenseNet with $1\times1$ conv integration layer is 3.51 \%, which is better than the error rate of 3.74 \% obtained by the original DenseNet ($k=24$).
However, the number of parameters of the WSMS-DenseNet increased from 27.2M to 28.0M.
For fair comparison, we also evaluated DenseNet ($k=26$).
Despite the increase in the number of parameters, DenseNet ($k=26$) achieved a worse error rate of 3.82 \%.
These results demonstrate that an increase in the number of parameters has a limit in the improvement of classification accuracy.
In contrast, our proposed WSMS-Net enables the original DenseNet to achieve a better classification accuracy with only a minor increase in number of parameters.
With WSMS-DenseNet, $1\times1$ conv and $3\times3$ conv integration layers achieved better results but no conv got worse results compared with the original DenseNet.
Recall that $1\times1$ conv and $3\times3$ conv can build a non-linear function, but no conv works just like an ensemble of features.
This difference demonstrates the importance of feature selection by the integration layer.
Between the two convolution methods of the integration layer, $1\times1$ conv achieved better results than $3\times3$ conv.
This result demonstrates that spatial convolution is unnecessary to select feature and $3\times3$ conv could be redundant in this respect.
Therefore, $1\times1$ conv is sufficient as the integration layer.
The MS-DenseNet achieved an error rate worse than those of the original DenseNet and WSMS-DenseNet despite the massive increase in the number of parameters, indicating that the improvement of WSMS-DenseNet is thanks to the shared parameters rather than the network architecture.
Even though the WSMS-DenseNet has a smaller increase in the number of parameters than the MS-Densenet thanks to the shared weight parameters, both the WSMS-DenseNet and MS-DenseNet have an equal increase in the number of calculations owing to the second and third stages.
Therefore, we calculated the number of multiplications over all the convolution layers in the original DenseNet and WSMS-DenseNet.
The number of multiplications is about 6,889M for the original 100-layer DenseNet and 8,454M for the 101-layer WSMS-DenseNet with a $1\times1$ conv integration layer: Our proposed WSMS-DenseNet incurs only a 20 \% increase in the number of multiplications.
The second stage requires less than 25 \% of the computations of the first stage because the input image is half the size.
In addition, for CIFAR-100, our proposed WSMS-DenseNet with the $1\times1$ conv integration layer achieved the lowest error rate of 18.45 \%.
DenseNet ($k=26$) and the MS-DenseNet also achieved accuracies better than that of the original DenseNet ($k=24$) but worse than that of the WSMS-DenseNet.
Next, we focus on the results of WSMS-DenseNet with 2 stages.
In the case of $1\times1$ conv and $3\times3$ conv, WSMS-DenseNet with $2$ stages on the CIFAR-10 achieved an accuracy worse than that with $3$ stages but better than the original DenseNet.
Decreasing the number of stages results in the lower robustness to scaling and accuracy improvement.
In the case of no conv, WSMS-DenseNet with $2$ stages achieved an accuracy worse than the original but better than WSMS-DenseNet with $3$ stages.
Without the feature selection by the extra convolution layer, a large number of stages is harmful to the classification accuracy.\\
\indent
For CIFAR-100, the results show a similar tendency to the case of the CIFAR-10.
WSMS-DenseNet with $2$ stages achieved a better accuracy than the original DenseNet only in the case of $3\times3$ conv.
We suppose that this is due to the complexity of the CIFAR-100.
Since the CIFAR-100 consists of a larger number of classes thus limited number of samples per class,
WSMS-Net has a much smaller chance of finding similar features in different stages from the CIFAR-100 images than from the CIFAR-10 images.
Hence, WSMS-Net requires more stages to learn similar features in different stages.
In the case of $3\times3$ conv, WSMS-DenseNet achieved a better accuracy simply thanks to the larger number of weight parameters like deeper DenseNet and MS-DenseNet.\\

\paragraph*{\textbf{Misclassified Images}}

In this section, we qualitatively evaluate our proposed WSMS-Net.
We collected test images from CIFAR-10 that DenseNet ($k=24$) and DenseNet ($k=26$) misclassified but WSMS-DenseNet ($k=24$) classified correctly in Fig.~\ref{fig:ScaleResult24}.
Many of them are close-ups and wide-shots.
We also collected CIFAR-10 test images randomly in Fig.~\ref{fig:CIFAR10TEST}, only a few of which are close-ups and wide-shots.
This difference demonstrates that WSMS-DenseNet ($k=24$) is more robust to scaling than the original DenseNet ($k=24$) and DenseNet ($k=26$).
As a conclusion, the WSMS-Net has become robust to object scaling, and therefore, it achieves a better classification accuracy than the original CNNs with only a limited increase in the number of weight parameters.

\begin{figure*}[!t]
\centering
\begin{tabular}{cc}
\includegraphics[width=3.8in,bb= 0 0 820 600,clip]{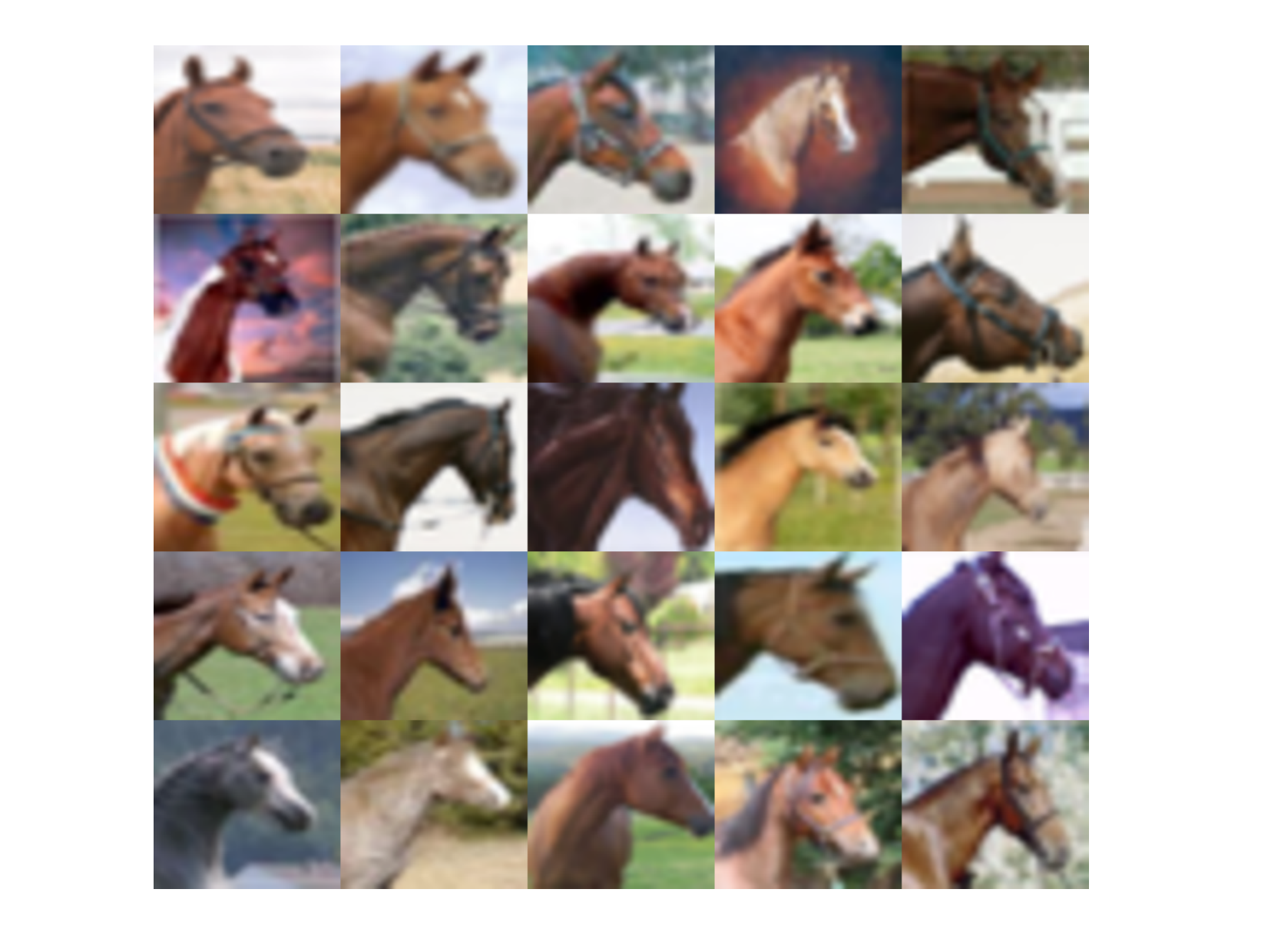}&
\includegraphics[width=3.8in,bb= 0 0 820 600,clip]{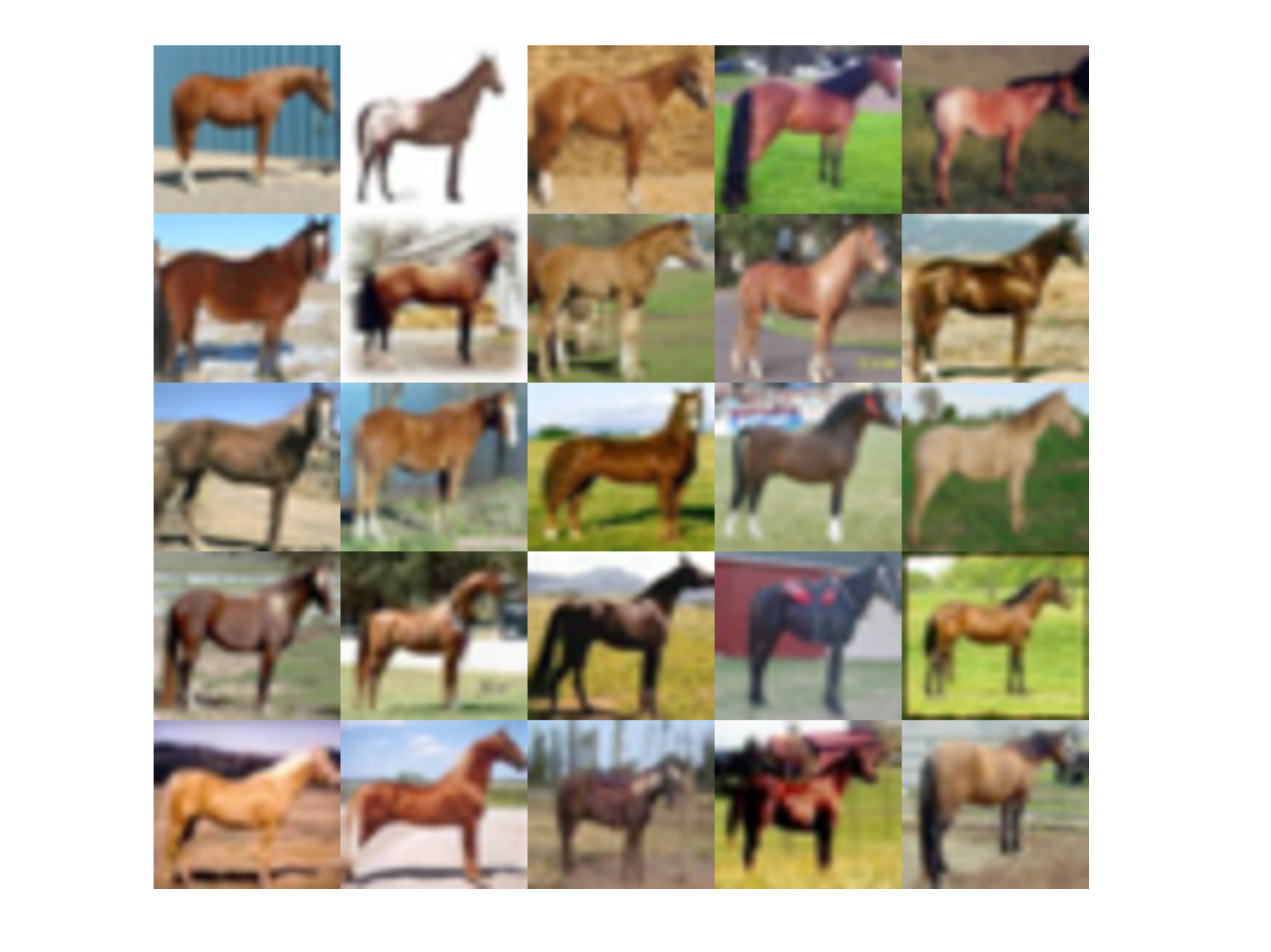}\\
\raisebox{1.5em}{(a)} & \raisebox{1.5em}{(b)} \\
\includegraphics[width=3.8in,bb= 0 40 820 570,clip]{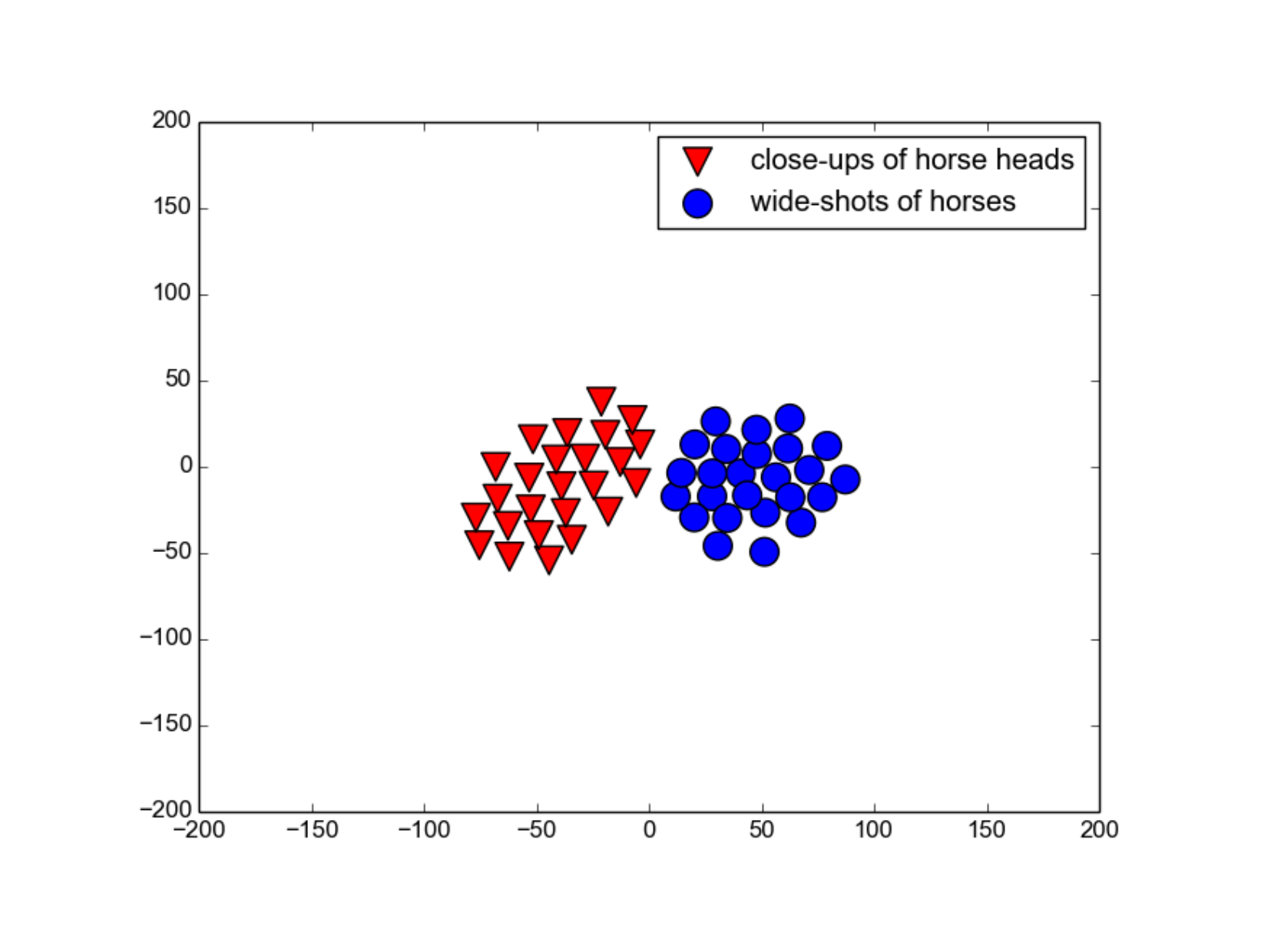}&
\includegraphics[width=3.8in,bb= 0 40 820 570,clip]{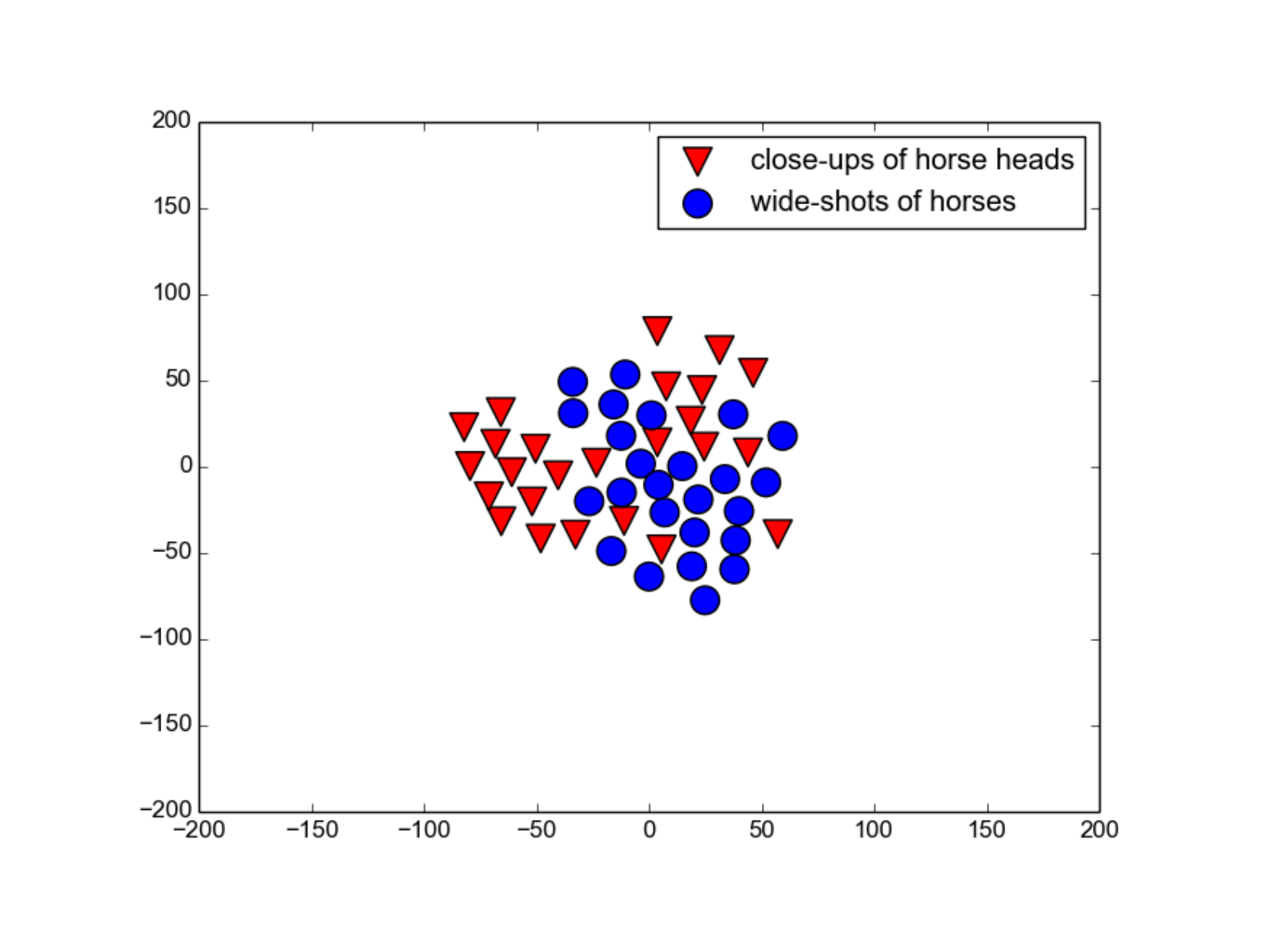}\\
\raisebox{1.5em}{(c)} & \raisebox{1.5em}{(d)} \\
\end{tabular}
\caption{
A $2$-dimensional visualization of the feature maps in the WSMS-DenseNet ($k=24$) before and after the integration layer.
We input two types of scaled horse images from CIFAR-10 test images; (a) close-ups of horse heads and (b) wide-shots of horses.
We used $8\times8$ average pooling and the t-SNE dimension reduction~\cite{Maaten2008} to obtain $2$-dimensional features for visualization.
(c) A visualization of feature maps of before the integration layer.
(d) A visualization of feature maps of after the integration layer.
}
\label{fig:tsne-result}
\end{figure*}

\paragraph*{\textbf{Visualization of Feature Selection by the Integration Layer}}
In Section~\ref{sec:integration}, we expected the integration layer to select the common feature (like red shaded areas in Fig.~\ref{fig:scale}) from the feature vectors obtained from multi-stages.
Here, we evaluate the feature selection.
We chose $50$ horse images from CIFAR-10 test images and fed them to the trained WSMS-DenseNet ($k=24$).
$25$ out of $50$ images are close-ups of horse heads and the remaining are wide-shots of horses as shown in Figs.~\ref{fig:tsne-result} (a) and (b).
We obtained their feature maps before and after the integration layer.
We performed $8\times8$ average pooling to these feature maps, resulting in $4,656$ and $128$-dimensional feature vectors.
We performed t-SNE dimension reduction~\cite{Maaten2008} to visualize these feature vectors in a $2$-dimensional space.
Fig.~\ref{fig:tsne-result} (c) shows that the feature vector before the integration layer forms two clusters depending on scales; Fig.~\ref{fig:tsne-result} (d) shows that the feature vector after the integration layer forms just one cluster.
This difference demonstrates that the integration layer selects features independent from the scale.


\subsection{Classification of CIFAR-10 and CIFAR-100 by WSMS-Pre-ResNet}

\paragraph*{\textbf{WSMS-Pre-ResNet on the CIFAR-10 and CIFAR-100}}

This section evaluates the WSMS-Pre-ResNet on the CIFAR-10 and CIFAR-100 datasets~\cite{Krizhevsky2009}.
The 1001-layer Pre-ResNet was reported to be the model achieving the highest accuracy for CIFAR-10 when ~\cite{He2016b} was published.
In contrast to a regular ResNet, the 1001-layer Pre-ResNet has a bottleneck architecture.
The 1001-layer Pre-ResNet has three compartments.
To downsample the feature map at the entrance of the second and third compartments, the 1001-layer Pre-ResNet sets the stride of the first of the three sequential convolution layers to 2.
The sizes of the feature maps are $32\times32$, $16\times16$, and $8\times8$ in the first, second, and third compartments, respectively.
The number of channels of the feature map is 3 at the input and is increased to 64 by the convolution layers before the first compartment.
After that, the feature map size is increased to 128, then 256, by the first convolution layers of the second and third compartments, respectively.
After the third compartment, global pooling is performed on the $8\times8$ feature map of $256$ channels, resulting in a $1\times1$ feature map of $256$ channels, i.e., a 256-dimensional feature vector.
This feature vector is given to the final fully connected layer for classification.

\begin{table*}[!t]
\renewcommand{\arraystretch}{1.3}
\caption{Test Error Rates of the WSMS-Pre-ResNet and Pre-ResNet on CIFAR-10/CIFAR-100.}
\centering
\scalebox{1.0}{
\label{tab:WSMS-ResNet_CIFAR-10}
\begin{tabular}{lccccc}
  \toprule
    & & \multicolumn{2}{c}{C10+} & \multicolumn{2}{c}{C100+} \\
  \cmidrule(l{.75em}r{.75em}){3-4}
  \cmidrule(l{.75em}r{.75em}){5-6}
    Network & depth & \#params & Error(\%) & \#params & Error(\%) \\
  \midrule
    Pre-ResNet                        & 1001  & 10.2M    & 4.62$^\ast$ & 10.2M & 22.71$^\dag$ \\
    Pre-ResNet                        & 1019  & 10.5M    & 4.62 & 10.5M & \textbf{21.70} \\
    MS-Pre-ResNet ($3$ stages, $1\times1$ conv) (not weight-shared for comparison)   & 1001  & 13.3M    & \textbf{4.01} & 13.3M & \textbf{20.09} \\
  \midrule
    WSMS-Pre-ResNet ($2$ stages, no conv)         & 1002  & 10.4M    & \textbf{4.08} & 10.4M & \textbf{20.28} \\
    WSMS-Pre-ResNet ($2$ stages, $3\times3$ conv) & 1002  & 10.8M    & \textbf{4.07} & 10.8M & \textbf{20.45} \\
    WSMS-Pre-ResNet ($2$ stages, $1\times1$ conv) & 1002  & 10.4M    & \textbf{3.97} & 10.4M & \textbf{20.32} \\
  \midrule
    WSMS-Pre-ResNet ($3$ stages, no conv)         & 1002  & 10.4M    & \textbf{4.05} & 10.4M & \textcolor{blue}{\textbf{19.83}} \\
    WSMS-Pre-ResNet ($3$ stages, $3\times3$ conv) & 1002  & 10.9M    & \textbf{3.95} & 10.9M & \textbf{20.39} \\
    \textbf{WSMS-Pre-ResNet ($3$ stages, $1\times1$ conv)} & 1002  & 10.4M    & \textcolor{blue}{\textbf{3.94}} & 10.4M & \textbf{20.49} \\
  \bottomrule
  \\
  \multicolumn{3}{l}{$^\ast$ indicates the result cited from original study~\cite{He2016b} and $^\dag$ indicates the result from our experiments.}
\end{tabular}
}
\end{table*}

We constructed the WSMS-Pre-ResNet by combining the 1001-layer Pre-ResNet with WSMS-Net.
We constructed versions of WSMS-Pre-ResNet with 2 and 3 stages so that each compartment of the 1001-layer Pre-ResNet corresponds to a convolution block of WSMS-Net.
A $32\times32$ input image was downsampled to $16\times16$ for the second stage and was downsampled to $8\times8$ for the third stage by a $2\times2$ average pooling layers.
The first stage was the same as the original 1001-layer Pre-ResNet before the global pooling layer.
The second stage was composed of the first two compartments, in which the size of the feature map was $16\times16$ in the first convolution block and $8\times8$ in the second convolution block.
In addition, the third stage was composed of the first compartment of the original 1001-layer Pre-ResNet using the $8\times8$ feature map.
The integration layer was given a feature map of $256+128+64=448$ channels.
We set $c$, the number of channels of the final feature map to 128, as is the case with the WSMS-DenseNet.
The hyperparameters and other structural parameters of the WSMS-Pre-ResNet followed those used for the 1001-layer Pre-ResNet in the original study~\cite{He2016b}.
Data normalization and data augmentation were performed in the same way as for the WSMS-DenseNet experiments.
Batch normalization~\cite{Ioffe2015} and ReLU activation function~\cite{Nair2010} were used.
The weight parameters were initialized following the algorithm proposed in ~\cite{He2016a}.
The WSMS-Pre-ResNet was trained using the momentum SGD algorithm with a momentum parameter of 0.9, mini-batch size of 64, and weight decay of $10^{-4}$ over 200 epochs.
The learning rate was initialized to 0.1, and then, it was changed to 0.01, and 0.001 at the 82nd and 123rd epochs, respectively.

\paragraph*{\textbf{Classification Results}}

Table~\ref{tab:WSMS-ResNet_CIFAR-10} summarizes the results of the WSMS-Pre-ResNet and original Rre-ResNet.
Following the original study~\cite{He2016b}, we obtained the mean test error rate of five trials.
We focus on the results of WSMS-Pre-ResNet with $3$ stages at first.
For CIFAR-10, the error rate of our proposed WSMS-Pre-ResNet with the $1\times1$ conv integration layer is 3.94 \%: This accuracy is clearly superior to the error rate of 4.62 \% obtained from the original Pre-ResNet.
We also evaluated a 1,019-layer Pre-ResNet for fair comparison and confirmed that it achieved the accuracies at the same level as the 1,001-layer Pre-ResNet despite a large increase in the number of weight parameters.
The MS-Pre-ResNet achieved an error rate better than the original Pre-ResNet but worse than the WSMS-Pre-ResNet despite the massive increase in the number of parameters.
The number of multiplications required by the original 1,001-layer Pre-ResNet is about 4,246M and that by the 1,002-layer WSMS-Pre-ResNet with $1\times1$ conv integration layer is about 5,045M: Our proposed WSMS-Pre-ResNet incurs only a 20 \% increase in the number of multiplications.

In the case of CIFAR-100, the performance of WSMS-Pre-ResNet with $1\times1$ conv integration layer also surpassed that of the original Pre-ResNet.
Unlike the case of CIFAR-10, the 1,019-layer Pre-ResNet and MS-Pre-ResNet achieved better results than the original Pre-ResNet.
We consider that the difference between CIFAR-10 and CIFAR-100 is caused by the complexity of the classification task.
Classification in CIFAR-100 is more difficult because of the larger number of classes and limited numbers of samples, and thus, requires much more weight parameters than classification in CIFAR-10.
Therefore, the 1,002-layer WSMS-Pre-ResNet and MS-Pre-ResNet, which have more weight parameters, are simply more advantageous than the original Pre-ResNet.
The original DenseNet already has enough parameters for classification of CIFAR-10 and CIFAR-100, but the original 1,001-layer Pre-ResNet does not have enough parameters for CIFAR-100 over 27.2M compared to 10.2M.
With WSMS-Pre-ResNet, all of $1\times1$ conv, $3\times3$ conv integration layer and no conv achieved better results than the original Pre-ResNet.
$1\times1$ conv achieved the highest accuracy on the CIFAR-10 as is the case with DenseNet.
However, no conv achieved the highest accuracy on the CIFAR-100.
We consider that this is due to the small number of channels of feature map: Pre-ResNet has only $256+128+64=448$ channels compared to the $2,320+1,552+784=4,656$ channels in DenseNet.
Because of this, the integration layer does not need to select the useful features, and the better accuracy is recorded by inputting all features to the last fully connected layer.
Especially in the case of CIFAR-100, due to its complexity of the classification, WSMS-Pre-ResNet needed the all channels of feature map to acquire as many clues as possible for classification.
WSMS-Pre-ResNet with $2$ stages achieved an accuracy almost comparable to but slightly worse than the one with $3$ stages in almost all the cases.
As expected, the reduced number of stages resulted in a worse accuracy because of the limited robustness to scaling.

Nonetheless, our proposed WSMS-Net with $1\times1$ conv enables the original Pre-ResNet to achieve better classification accuracy with only limited increase in the number of parameters, even for CIFAR-100.


\begin{table*}[!t]
\renewcommand{\arraystretch}{1.3}
\caption{Test Error Rates of the DenseNet and Pre-ResNet with image scaling data augmentation, multi-scale DenseNet and multi-scale Pre-ResNet on the CIFAR-10/CIFAR-100.}
\centering
\scalebox{1.0}{
\label{tab:Compare}
\begin{tabular}{lcc}
  \toprule
    Network & CIFAR-10 Error(\%) & CIFAR-100 Error(\%) \\
  \midrule
    DenseNet                                                  & 3.74$^\ast$  & 19.25$^\dag$ \\
    DenseNet (data augmentation, only $32\times32$ test data) & 11.87 & 46.03 \\
    DenseNet (data augmentation, ensemble of sizes of $8$, $16$ and $32$) \cite{Marmanis2016a}                   & 6.21  & 28.73 \\
    multi-scale DenseNet ~\cite{Cai2016}             & 5.04  & 23.55 \\
    \textbf{WSMS-DenseNet ($3$ stages, $1\times1$ conv)}                           & \textcolor{blue}{\textbf{3.51}}  & \textcolor{blue}{\textbf{18.45}} \\
  \midrule
    Pre-ResNet                                                  & 4.62$^\ast$  & 22.71$^\ast$ \\
    Pre-ResNet (data augmentation, only $32\times32$ test data) & 13.61 & 52.86 \\
    Pre-ResNet (data augmentation, ensemble of sizes of $8$, $16$ and $32$) \cite{Marmanis2016a}                 & 9.23  & 35.47 \\
    multi-scale Pre-ResNet ~\cite{Cai2016}             & 5.38  & \textbf{21.87} \\
    \textbf{WSMS-Pre-RseNet ($3$ stages, $1\times1$ conv)}                           & \textcolor{blue}{\textbf{3.94}}  & \textcolor{blue}{\textbf{20.49}} \\
  \bottomrule
  \multicolumn{3}{l}{$^\ast$ indicates the result cited from original studies~\cite{Huang2016b,He2016b} and $^\dag$ indicates the result from our experiments.}
\end{tabular}
}
\end{table*}

\subsection{Comparison of WSMS-Net and image scaling data augmentation}
\label{sec:Dataaug}
\label{com1}
In this section, we compare the performance of our WSMS-Net and the original networks with data augmentation which resizes the input image.
This is called the \emph{image scaling data augmentation}, hereafter.
We used the DenseNet and Pre-ResNet as the original networks of our WSMS-Net and the image scaling data augmentation.
We evaluated WSMS-DenseNet, WSMS-Pre-ResNet, DenseNet with image scaling data augmentation, and Pre-ResNet with image scaling data augmentation on the CIFAR-10 and CIFAR-100 datasets.
\paragraph*{\textbf{The Image Scaling Data Augmentation}}
For DenseNet and Pre-ResNet with the image scaling data augmentation, we prepared $16\times16$ and $8\times8$ images resized from the original $32\times32$ images additionally.
The sizes of images were randomly determined in every training iteration.
In the test phase, we evaluated the networks using the following images.%
\begin{enumerate}
  \item images of the original size as usual.
  \item images of the original, $\frac{1}{2}$, and $\frac{1}{4}$ size with an ensemble.
\end{enumerate}
The ensemble method followed Marmanis et al. \cite{Marmanis2016a}, which used an ensemble of multiple outputs corresponding to an input image of multiple scales.
The hyperparameters and other conditions were the same as those used in the original studies of Pre-ResNet and DenseNet.
\paragraph*{\textbf{Classification Results}}
Table~\ref{tab:Compare} shows the results of the DenseNet and Pre-ResNet with the image scaling data augmentation as well as the WSMS-Net.
The DenseNet and Pre-ResNet with the image scaling data augmentation achieved the worse accuracies.
The ensemble improved the accuracies, but they were still worse than the originals.
These results indicate that the DenseNet and Pre-ResNet tried to learn the multiple features from an input of multiple scales, but they do not have enough capacity (i.e., weight parameters) to do that straight-forwardly as mentioned in Section~\ref{sec:concept}.
We conclude that out WSMS-Net is different than and superior to the image scaling data augmentation.

\subsection{Comparison with Existing CNNs using Multi-scale Feature Fusion}
\label{sec:previous_work}
\label{com2}
\paragraph*{\textbf{Multi-Scale CNN for image classification}}
In this section, we compare our WSMS-Net with a previous work focusing on dealing with scaling, the multi-scale CNN~\cite{Cai2016}, mentioned in Section \ref{sec:Related_works}.
The multi-scale CNN focuses on object detection task originally but it is a multi-scale feature fusion method for the robustness to scaling like WSMS-Net.
Multi-scale CNN has multiple output layers (one from the last layer and the others from intermediate layers just before downsampling layers) so that their receptive fields correspond to objects of different scales.
In other words, multi-scale CNN uses intermediate feature maps like the second and following stages of WSMS-Net.
For the comparison with our WSMS-Net, we applied the architecture of multi-scale CNN to the DenseNet and Pre-ResNet and evaluated the performance of classification on the CIFAR-10 and CIFAR-100.
We call DenseNet and Pre-ResNet with multi-scale CNN as the ``multi-scale DenseNet'' and ``multi-scale Pre-ResNet'', respectively.
The original DenseNet and Pre-ResNet have three conv blocks, and hence, we added additional output layers to the last layers of the first and second conv blocks.
Following the original multi-scale CNN, the total loss function was the sum of the loss functions of the three output layers.
The hyperparameters and other conditions were the same as those used in the original studies of Pre-ResNet, DenseNet, and multi-scale CNN.
\paragraph*{\textbf{Classification Results}}
Table~\ref{tab:Compare} shows the results of the multi-scale DenseNet and multi-scale Pre-ResNet.
Both of them resulted the worse accuracies except for the multi-scale Pre-ResNet on CIFAR-100.
These results indicate that the multi-scale CNN either does not cooperate with general deep CNNs like DenseNet and Pre-ResNet, does not work well for the classification task, or at least requires a special adjustment of hyperparameters.
According to these results and the results in Section \ref{sec:Dataaug}, we conclude that only one of resizing input and taking multiple intermediate features does not work well.
Our WSMS-Net improved classification accuracy by resizing input, taking multiple intermediate features, and then, selecting features at the integration layer.


\begin{table*}[!t]
\renewcommand{\arraystretch}{1.3}
\caption{Single Crop Test Error Rates of WSMS-ResNets and Original ResNets on ImageNet Dataset.}
\centering
\label{tab:WSMS-ResNet_ImageNet}
\begin{tabular}{lcccc}
  \toprule
    Network                               & depth & \#params & top-1 Error(\%) & top-5 Error(\%)  \\
  \midrule
    ResNet                                &   50   & 25.6M   & 24.01           & 7.02             \\
    WSMS-ResNet ($3$ stages, $1\times1$ conv)  &   51   & 28.3M   & \textbf{23.07}  & \textbf{6.44}    \\
    WSMS-ResNet ($4$ stages, $1\times1$ conv)  &   51   & 28.5M   & \textcolor{blue}{\textbf{23.13}}  & \textcolor{blue}{\textbf{6.57}}    \\
  \midrule
    ResNet                                &  101   & 44.5M   & 22.44           & 6.21             \\
    WSMS-ResNet ($3$ stages, $1\times1$ conv)  &  102   & 47.3M   & \textbf{22.20}  & 6.22             \\
    WSMS-ResNet ($4$ stages, $1\times1$ conv)  &  102   & 47.6M   & \textcolor{blue}{\textbf{22.09}}  & \textcolor{blue}{\textbf{6.06}}             \\
  \midrule
    ResNet                                &  152   & 60.2M   & 22.16           & 6.16             \\
    WSMS-ResNet ($3$ stages, $1\times1$ conv)  &  153   & 63.0M   & \textbf{21.99}  & \textbf{6.04}    \\
    WSMS-ResNet ($4$ stages, $1\times1$ conv)  &  153   & 63.3M   & \textcolor{blue}{\textbf{21.93}}  & \textcolor{blue}{\textbf{5.90}}    \\
  \bottomrule
\end{tabular}
\\[1mm]
\begin{minipage}{11.5cm}
We omitted the experiments of the WSMS-ResNet with $2$ stages because of the worse results of the WSMS-Pre-ResNet and WSMS-DenseNet with $2$ stages than with $3$ stages on the CIFAR dataset. (see Tables~\ref{tab:WSMS-DenseNet_CIFAR-10} and~\ref{tab:WSMS-ResNet_CIFAR-10})
\end{minipage}
\end{table*}

\subsection{ImageNet classification by WSMS-ResNet}

\paragraph*{\textbf{WSMS-ResNet on the ImageNet}}
This section evaluates ResNet~\cite{He2016} with WSMS-Net for the ImageNet 2012 classification dataset~\cite{Russakovsky2014}.
ImageNet consists of 1.28 million training images and 50,000 validation images.
Each image is given one of 1,000 class labels.
We used 50-layer, 101-layer and 152-layer ResNets with the bottleneck architecture as the base to construct our WSMS-Nets.
To downsample the feature map at the entrance of each compartment, a ResNet with bottleneck architecture modifies the stride of the second of the three sequential convolution layers to 2, and replaces the shortcut with a $1\times1$ convolution layer with a stride of 2, while the original ResNet employs pooling layers.

The ResNet is given an $N\times M$ RGB input image.
The input image becomes an $\frac{N}{4}\times\frac{M}{4}$ feature map of $64$ channels through a $7\times7$ convolution layer with a stride of $2$ and a following $3\times3$ max pooling layer with a stride of $2$.
The remaining part of ResNet has four compartments, each consisting of several residual blocks with bottleneck architecture.
The size of the feature maps are $\frac{N}{4}\times\frac{M}{4}$, $\frac{N}{8}\times\frac{M}{8}$, $\frac{N}{16}\times\frac{M}{16}$, and $\frac{N}{32}\times\frac{M}{32}$, and the number of channels of the feature maps are 256, 512, 1,024, and 2,048 for the first, second, third, and fourth compartments, respectively.
After the fourth compartment, global pooling is performed, resulting in a $1\times1$ feature map of 2,048 channels.
This feature vector is given to the final fully connected layer for 1,000-class classification.
The numbers of residual blocks are $3$, $4$, $6$, and $3$ in the first, second, third, and forth compartments of the 50-layer ResNet, $3$, $4$, $23$, and $3$ in the 101-layer ResNet, and $3$, $8$, $36$, and $3$ in the 152-layer ResNet, respectively.

We constructed the WSMS-ResNets by combining the 50-layer, 101-layer, and 152-layer ResNets~\cite{He2016} for ImageNet dataset with WSMS-Net.
The number of stages was set to three and four.
In the case of ImageNet experiment, the original ResNet has four compartments, and we can set the number of stages to two, three, or four.
We omitted the experiment of WSMS-ResNet with $2$ stages because of the worse results of WSMS-Pre-ResNet and WSMS-DenseNet with $2$ stages than with $3$ stages on the CIFAR-10 and CIFAR-100 shown in Tables~\ref{tab:WSMS-DenseNet_CIFAR-10} and~\ref{tab:WSMS-ResNet_CIFAR-10}.
The first stage was just the same as the original ResNet before the global pooling layer.
An $N\times M$ input image was downsampled to $\frac{N}{2}\times\frac{M}{2}$ for the second stage and $\frac{N}{4}\times\frac{M}{4}$ for the third stage.
The second stage was composed of the first three compartments of the original ResNet, in which the size of the feature map was $\frac{N}{8}\times\frac{M}{8}$, $\frac{N}{16}\times\frac{M}{16}$, and $\frac{N}{32}\times\frac{M}{32}$ in the first, second, and third blocks, respectively.
In addition, the third stage was composed of the first two compartments, using $\frac{N}{16}\times\frac{M}{16}$ and $\frac{N}{32}\times\frac{M}{32}$ feature maps.
The integration layer was given a $\frac{N}{32}\times\frac{M}{32}$ feature map of $2,048+1,024+512=3,584$ channels.
We set $c=1,024$.
We evaluated our WSMS-ResNets only with the $1\times1$ conv integration layer because of the results in the previous sections.

The hyperparameters and other conditions of WSMS-ResNets followed those used in the original study~\cite{He2016} and the reimplementation by Facebook AI Research posted on \url{https://github.com/facebook/fb.resnet.torch}.
Batch normalization~\cite{Ioffe2015} and a ReLU activation function~\cite{Nair2010} were used.
The weight parameters were initialized following the algorithm proposed in ~\cite{He2016a}.
The WSMS-ResNets were trained using the momentum SGD algorithm with a momentum parameter of 0.9, mini-batch size of 256, and weight decay of $10^{-4}$ over 90 epochs.
The learning rate was initialized to 0.1, and then, it was reduced to 0.01 and 0.001 at the 30th and 60th epochs, respectively.

\paragraph*{\textbf{Classification Results}}
Table~\ref{tab:WSMS-ResNet_ImageNet} summarizes the results of the WSMS-ResNets and original ResNets for ImageNet classification.
The classification results of the original ResNets were cited not from the original study~\cite{He2016} but the reimplementation by Facebook AI Research posted on \url{https://github.com/facebook/fb.resnet.torch}.
Note that Facebook's results are better than those shown in the original study~\cite{He2016}.

We first focus on the results of WSMS-ResNet with $4$ stages.
According to Table~\ref{tab:WSMS-ResNet_ImageNet}, our proposed 51-layer, 102-layer, and 153-layer WSMS-ResNets achieved better results than their original counterparts, in both top-1 and top-5 error rates.
While the WSMS-ResNets for ImageNet classification have more parameters than the original ResNets, the 102-layer WSMS-ResNet with $4$ stages having $47.6$M parameters is superior to the original 152-layer ResNet having $60.2$M parameters.
This is evidence that our proposed WSMS-Net improves the original CNN while maintaining a limited increase in the number of weight parameters.
WSMS-ResNet with $3$ stages achieved accuracies worse than the one with $4$ stages but better than the original ResNet, except for the 102-layer WSMS-ResNet, top-5 error rate.
This indicates that WSMS-Nets work better with even more stages.

All the results for the CIFAR-10, CIFAR-100, and ImageNet classification tasks demonstrate that a combination with our proposed WSMS-Net contributes to the various datasets and architectures of CNNs.

\section{Conclusion}

In this study, we proposed a novel network architecture for convolutional neural networks (CNNs) called the \emph{weight-shared multi-stage network} (WSMS-Net) to improve classification accuracy by acquiring robustness to object scaling.
The WSMS-Net consists of multiple stages of CNNs, each given input image of different sizes.
Feature maps obtained from all the stages are concatenated and integrated at the ends of the stages.
The increases in the number of weight parameters and computations were limited.
The experimental results demonstrated that the WSMS-Net achieved better classification accuracy and had the higher robustness to object scaling than existing CNN models while deepening the network is a less parameter-efficient way to improve classification accuracy.
Future works include a more detailed evaluation of the robustness to object scaling, evaluation on additional datasets, and evaluation with other CNN architectures.

\section*{Acknowledgment}
This study was partially supported by the MIC/SCOPE \#172107101.



\begin{IEEEbiography}[{\includegraphics[width=1in,height=1.25in,bb= 0 0 596 842,clip,keepaspectratio]{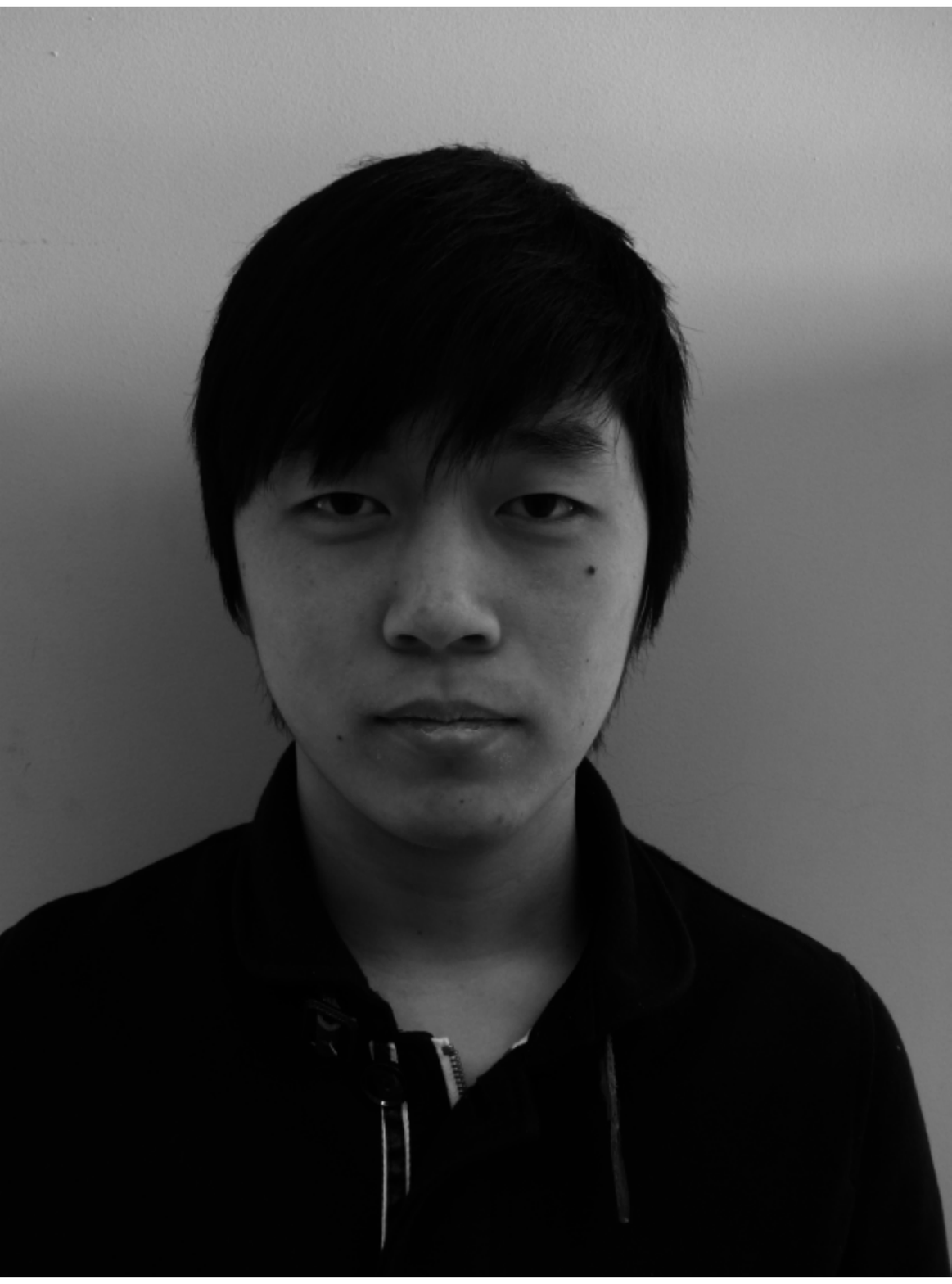}}]%
{Ryo Takahashi}
is a graduate student of Graduate School of System Informatics, Kobe University, Hyogo, Japan. He received the B.E.~degree from Kobe University. He investigated image classification by deep neural network architectures.
\end{IEEEbiography}

\begin{IEEEbiography}[{\includegraphics[width=1in,height=1.25in,bb= 0 0 360 480,clip,keepaspectratio]{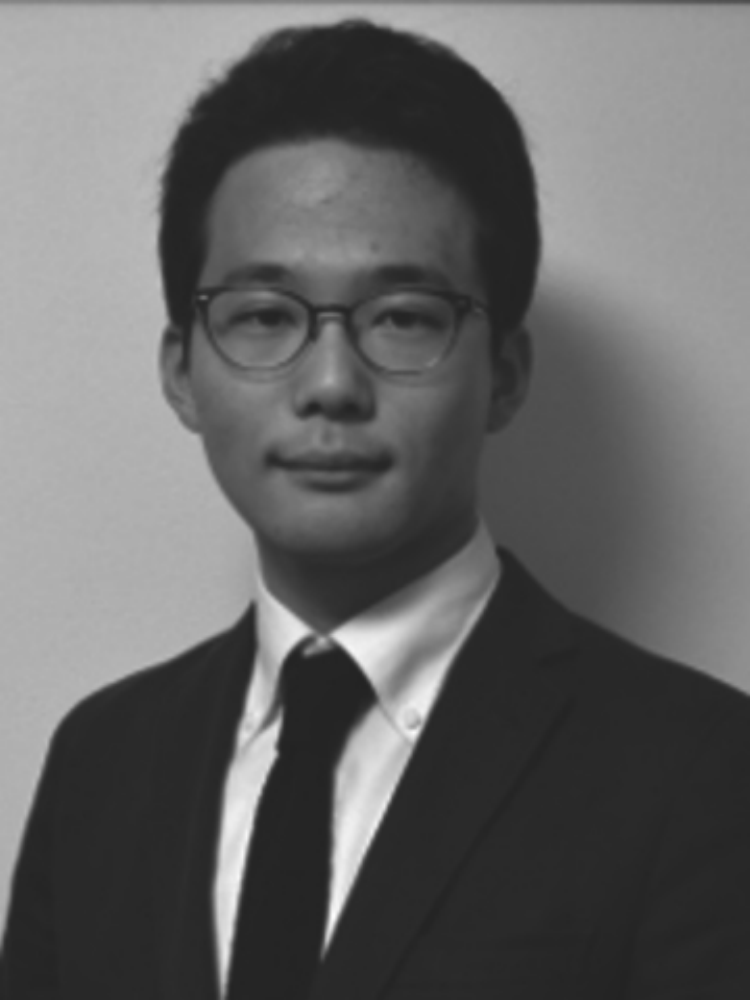}}]%
{Takashi Matsubara}
received his B.E., M.E., and Ph.D.~in engineering degrees from Osaka University, Osaka, Japan, in 2011, 2013, and 2015, respectively.
He is currently an assistant professor at the Graduate School of System Informatics, Kobe University, Hyogo, Japan. His research interests are in computational intelligence and computational neuroscience.
\end{IEEEbiography}

\begin{IEEEbiography}[{\includegraphics[width=1in,height=1.25in,bb=0 0 162 180,clip,keepaspectratio]{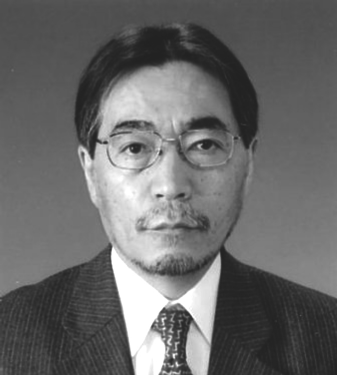}}]%
{Kuniaki Uehara}
received his B.E., M.E., and D.E.~degrees in information and computer sciences from Osaka University, Osaka, Japan, in 1978, 1980 and 1984, respectively. From 1984 to 1990, he was with the Institute of Scientific and Industrial Research, Osaka University as an Assistant Professor. From 1990 to 1997, he was an Associate Professor with Department of Computer and Systems Engineering of Kobe University. From 1997 to 2002, he was a Professor with the Research Center for Urban Safety and Security of Kobe University. Currently he is a Professor with Graduate School of System Informatics of Kobe University.
\end{IEEEbiography}

\end{document}